%% file: main.tex
\documentclass[a4paper,twoside]{article}

\usepackage{epsfig}
\usepackage{subcaption}
\usepackage{calc}
\usepackage{amssymb}
\usepackage{amstext}
\usepackage{amsmath}
\usepackage{amsthm}
\usepackage{multicol}
\usepackage{pslatex}
\usepackage{apalike}
\usepackage{algorithm2e}
\usepackage[bottom]{footmisc}
\usepackage{cite}

\usepackage{siunitx} 
\usepackage{hyperref}
\hypersetup{colorlinks}
\usepackage{subcaption}
\usepackage{cleveref}
\usepackage{comment}
\usepackage{float}
\usepackage{tikz}
\usepackage{pgfplots}
\usepackage[inline]{enumitem}
\usepackage{SCITEPRESS}     

\begin{document}

\title{Adaptive Prompt Tuning: Vision Guided Prompt Tuning with Cross-Attention for Fine-Grained Few-Shot Learning}


\author{\authorname{Eric Brouwer \sup{1,2}, Jan Erik van Woerden\sup{2}, Gertjan Burghouts\sup{2},
Matias Valdenegro-Toro\sup{1} and Marco Zullich\sup{1}}
\affiliation{\sup{1}Faculty of Science and Engineering, University of Groningen, Nijenborgh 9, 9747 AG, Groningen, the Netherlands}\affiliation{\sup{2}TNO, Oude Waalsdorperweg 63, 2597 AK, Den Haag, the Netherlands}
\email{ericbrouwer0@gmail.com, jan\_erik.vanwoerden@tno.nl, gertjan.burghouts@tno.nl, m.a.valdenegro.toro@rug.nl, marco.zullich@gmail.com}
}

\keywords{CLIP, visual prompt tuning, Few-Shot learning, Fine-grained image recognition, adaptive inference, Uncertainty Quantification, Monte-Carlo dropout, expected calibration error }

\abstract{
Few-shot, fine-grained classification in computer vision poses significant challenges due to the need to differentiate subtle class distinctions with limited data. This paper presents a novel method that enhances the Contrastive Language-Image Pre-Training (CLIP) model through adaptive prompt tuning, guided by real-time visual inputs. Unlike existing techniques such as Context Optimization (CoOp) and Visual Prompt Tuning (VPT), which are constrained by static prompts or visual token reliance, the proposed approach leverages a cross-attention mechanism to dynamically refine text prompts for the image at hand. This enables an image-specific alignment of textual features with image patches extracted from the Vision Transformer, making the model more effective for datasets with high intra-class variance and low inter-class differences. The method is evaluated on several datasets, including CUBirds, Oxford Flowers, and FGVC Aircraft, showing significant performance gains over static prompt tuning approaches. To ensure these performance gains translate into trustworthy predictions, we integrate Monte-Carlo Dropout in our approach to improve the reliability of the model predictions and uncertainty estimates. This integration provides valuable insights into the model's predictive confidence, helping to identify when predictions can be trusted and when additional verification is necessary. This dynamic approach offers a robust solution, advancing the state-of-the-art for few-shot fine-grained classification.}

\onecolumn \maketitle \normalsize \setcounter{footnote}{0} \vfill

\begin{figure*}[h]
    \centering
    \includegraphics[width=1\linewidth]{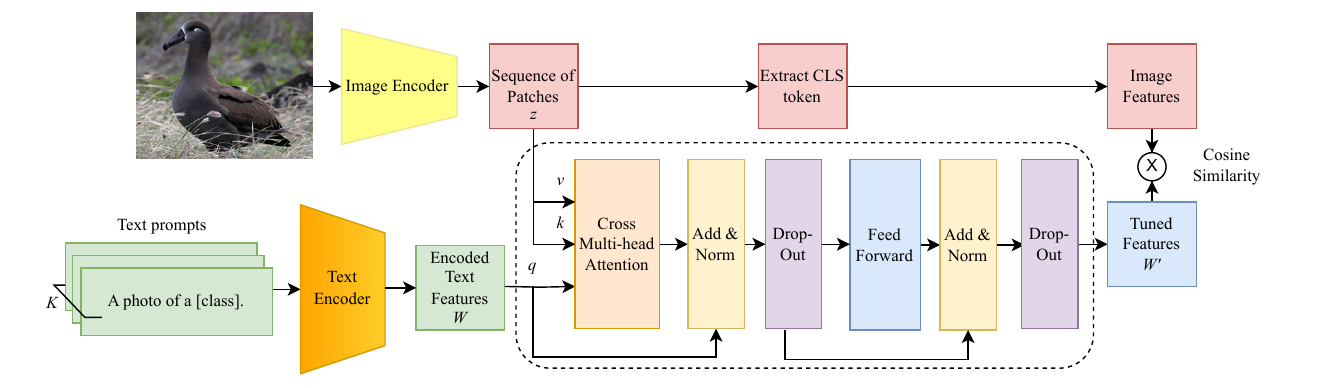}
    \caption{Overview of the proposed APT method. 
    The method leverages CLIP's image and text encoder (see \Cref{fig:clip_architecture}) to refine the text embeddings for the few-shot classification task.
    The main novelty introduced by APT is the cross-attention layer, illustrated within the dotted lines.
    It combines visual and text information, merging them thanks to the cross multi-head attention operation.
    The resulting output is passed through normalization, dropout, a feed forward layer---responsible for adding non-linearity to the process, and skip connections to produce a set of tuned features, which can netter fit the images at hand for performing the few-shot classification task.
    This layer is the only component which is trained in the few-shot problem---the weights of image and text encoders are frozen into their pretrained state.
    The tuned features are later related to the image features using the cosine similarity (see \Cref{eq:cosine_sim}) for operating the few-shot classification.
    }
    \label{fig:architecture}
\end{figure*}

\section{Introduction}
\label{sec:introduction}

In recent years, the field of computer vision has experienced remarkable growth and transformation \cite{alom2019state}, driven by significant advancements in machine learning and deep learning technologies. This progress is partly attributable to the development of large-scale pre-trained models, commonly referred to as \emph{foundation models}.
By training on extensive datasets containing millions of data, these models are able to perform exceptionally well even on previously unseen tasks.
Among these foundation models, CLIP (Contrastive Language-Image Pre-Training) \cite{radford2021learning} has emerged as a particularly influential tool. CLIP involves learning a joint embedding space for both textual and visual data using contrastive learning on a large corpus of text-image pairs.


Foundation models can be fine-tuned to solve specific \emph{downstream tasks} achieving state-of-the-art performance while often requiring fewer computational resources if compared to training a model from scratch \cite{jena2022high}.
This also translates to situations whenever data availability is limited, a situation commonly identified as \emph{zero}-- and \emph{few-shot learning} \cite{lemley2017deep}.
In these cases, Deep Neural Networks trained from scratch on these limited datasets have been shown to be severely overfitting \cite{nakkiran2021deep}.
CLIP embeddings can be used as-is (\emph{static prompting}) on the downstream task to perform \emph{zero-shot learning}; additionally, the embeddings can be adapted to a specific dataset (\emph{dynamic prompting}) to perform, e.g., \emph{few-shot learning}.

Previous works such as Context Optimization (CoOp) \cite{zhou2022conditional} and Visual Prompt Tuning (VPT) \cite{jia2022visual} have been explored to enhance CLIP for few-shot learning; however, they are still prone to poor generalization, especially on fine-grained classification tasks. These approaches augment the input to the model by adding a learnable context vector or visual prompts, which are trained on the limited few-shot dataset to improve performance. The learnable context often becomes specialized to these few examples and consequently, the model may overfit.
In response to these limitations, this paper proposes Adaptive Prompt Tuning (APT), a dynamic method that adapts prompts at inference time through a cross-attention mechanism \cite{vaswani2017attention} between image and text features.
This approach enhances CLIP’s ability to align textual and visual elements in fine-grained few-shot learning, allowing it to dynamically focus on relevant features.

In addition, we enhance APT by means of Monte-Carlo Dropout (MCD) in order to calibrate its output, showing improved confidence estimates without drops in accuracy.
Traditional Deep Neural Networks have been shown to consistently produce overconfident outputs \cite{nguyen2015deep}.
The usage of techniques to improve Uncertainty Quantification (UQ) allows models to output predictions whose confidence is comparable to their accuracy---a situation which is 
instrumental in critical applications, such as medical applications \cite{kim2022calibration}, thus allowing underconfident predictions---which are likely to be wrong---to be discarded, improving the reliability of the model.

On the topic of CLIP-based few-shot segmentation,  \cite{miao2024bayesian} are, to the best of our knowledge, the only authors proposing a method for estimating uncertainty on this topic.
They enriched the deterministic CLIP predictions with an ensemble of Gaussian Processes (GPs) to induce a probability distribution on the outputs to produce the uncertainty estimates.
Despite their model being effective, GPs are noticeable for being memory intensive, needing the whole training dataset for generating predictions.
Our method instead uses a \emph{classical} Neural Network--based approach which needs to store only the parameters of the cross-attention layer, regardless of the dataset size, to produce the uncertainty estimates.

In summary, the contribution of our papers are the following:
\begin{itemize}
    \item We introduce APT, a novel and competitive cross-attention-based approach for CLIP-guided fine-grained few-shot image classification, and
    \item We enhance APT using MCD, in order to produce an analysis on the uncertainty estimates output by APT.
\end{itemize}

Our code is available on GitHub: \url{https://github.com/ericbrouwer0/adaptive-prompt-tuning}.

\section{Related Works}

\paragraph{Vision-language Models} 

Vision-language models (VLMs) integrate computer vision and natural language processing to jointly learn representations of visual and textual data. 
By embedding images and textual descriptions into a shared space, VLMs enable tasks like zero-shot classification, image captioning, and visual search without the need for task-specific training data \cite{chen2020simple, henaff2020data}, leveraging the possibility of guiding the classification through natural-language text prompting.
CLIP \cite{radford2021learning}, specifically, has been shown to be an effective model at performing image recognition tasks in a zero-- of few-shot setting, accurately categorizing images into a wide variety of classes without direct exposure to those specific classes during training. CLIP's image encoder processes visual inputs into a high-dimensional feature space, while the text encoder similarly processes textual inputs. These two vectors are then projected into a shared embedding space, where they can be compared directly using a cosine similarity metric, enabling the model to make predictions based on textual descriptions of visual categories. 

\paragraph{Few-shot Learning}
Few-shot learning aims to enable models to recognize new tasks or objects with minimal data, inspired by human cognitive abilities to generalize from few examples \cite{wang2020generalizing}. 
Deep Neural Networks rely heavily on large datasets to achieve state-of-the-art performance; however, they tend to overfit on small datasets \cite{nakkiran2021deep}, thus often being unsuitable for few-shot scenarios.
Several methods have been proposed to address this challenge. Meta-learning trains models across various tasks to help them quickly adapt to new tasks with minimal data \cite{chen2021meta}. Prototypical networks \cite{snell2017prototypical} offer another solution by learning a metric space where classification is based on the distance to prototype representations of each class \cite{ding2020graph}.  
More recent approaches are based on fine-tuning large foundation models such as CLIP, leveraging 
\begin{enumerate*}[label=(\alph*)]
\item the generalization capabilities of models pre-trained on vast amounts of data and
\item the aforementioned possibility of using text prompts for guiding the classification.
\end{enumerate*}
Prompt tuning has emerged as a viable approach to fine-tuning foundation models for few-shot learning. Frameworks like CoOp \cite{zhou2022learning} enhance CLIP by learning task-specific prompt embeddings. Extensions like CoCoOp \cite{zhou2022conditional} further improve robustness to unseen classes by incorporating image features. Additionally, visual prompt tuning (VPT) \cite{jia2022visual}, which tunes image encoders with learnable task-specific prompts, has proven effective in low-data scenarios, preserving generalization while minimizing the need for extensive retraining or large labeled datasets.

\paragraph{Fine-grained Recognition}

Fine-grained image recognition focuses on the task of distinguishing between highly similar subcategories within a larger, general category, such as identifying specific species of birds \cite{WahCUB_200_2011}, types of cars \cite{dehghan2017view}, or types of air-crafts \cite{maji13fine-grained}. This domain presents a unique set of challenges that diverge significantly from those encountered in more generalized image classification tasks. The nuances and subtle differences that define each subcategory require models to develop a sense of discriminative feature detection, far beyond what is typically necessary for distinguishing between broadly defined classes \cite{peng2017object} such as a car and a person. \\

\paragraph{Uncertainty Quantification}

UQ plays a critical role in assessing the confidence of machine learning models, especially in high-stakes applications. By evaluating the reliability of predictions, UQ helps to identify areas where models may fail or need improvement, thus increasing the robustness and trustworthiness of AI systems.
Within the framework of Deep Learning, (approximate) Bayesian Neural Networks (BNNs) offer a strong framework for quantifying predictive uncertainty by placing distributions over model parameters rather than learning fixed weights \cite{goan2020bayesian}.
At inference time, a predictive distribution---rather than a point one---is produced, allowing for considerations on the predictive uncertainty. 
While exact Bayesian inference is often unfeasible to implement in deep learning, approximate methods, like MCD, Bayes-by-backprop \cite{blundell2015weight}, and Deep Ensembles \cite{ganaie2022ensemble} are often used instead.
Specifically, MCD \cite{gal2016dropout} operates by using dropout as a way of inducing stochasticity in the output.
Despite often showcasing worse UQ capability with regards to other tools, MCD is still used due to its simplicity, since it can be used straight away on architectures which already employ Dropout \cite{valdenegro2022deeper}.
Within the field of guided prompt tuning for fine-grained few-shot learning, we have only identified one work \cite{miao2024bayesian} applying a Bayesian framework, which augments the visual and text embeddings produced by CLIP, in order to produce uncertainty estimates.

\paragraph{Our contribution}

In summary, the contributions of the present work are as follows:
\begin{itemize}
    \item We introduce a novel cross-attention based prompt tuning mechanism that jointly optimizes visual and text embeddings, delivering a competitive or superior performance relative to the state-of-the-art guided prompt tuning approaches.
    \item We conduct a comprehensive UQ analysis, demonstrating notable improvements in the quality of uncertainty estimates produced by our proposed model.
\end{itemize}

\section{Materials and Methods}

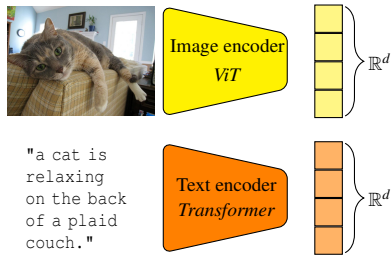
\begin{figure}
    \centering
    \input{figs/diagrams/clip_tikz}
    \caption{Diagram depicting the architecture of CLIP used in the present work.
    CLIP works by employing two Deep Neural Networks---an image encoder and a text encoder.
    The image encoder, a Vision Transformer (ViT), embeds the image into the $\mathbb{R}^d$ space; the text encoder embeds a natural language sentence in the same space.
    Image and text pair selected from the COCO dataset \cite{lin2014microsoft}.
    }
    \label{fig:clip_architecture}
\end{figure}

\subsection{CLIP}
The CLIP model uses a dual encoding mechanism, one for images and another for text, aimed at learning combined visual-textual representations. Each component is specialized to translate its respective input into a common feature space where the semantic contents of both modalities are directly comparable. \\
As represented in \Cref{fig:clip_architecture}, CLIP employs two Deep Neural Networks that are tasked with jointly encoding (image, text) pairs.
The CLIP model uses Transformer architectures for both image and text encoding, with the main difference being in the input processing. The image encoder (ViT) \cite{dosovitskiy2020image} splits the input image ($H,W,C$) into $16 \times 16$ patches and linearly embeds them, while the text encoder tokenizes and embeds the input text. Both encoders then use a series of self-attention layers to produce final encodings of the same dimension $d$. 
\cite{radford2021learning} trained CLIP in a self-supervised fashion on a large dataset of text-image pairs.
They employed contrastive learning with the goal of creating an embedding space in which positive, i.e., related, (image, text) pairs are pulled closer together, while negative, i.e., unrelated, pairs are pushed apart.

\paragraph{Zero-Shot inference with CLIP}

The setup for a classification task can be performed by taking the features generated by the image encoder, and using the cosine similarity metric to compare this to a set of encoded prompts that act as the relevant categories \cite{radford2021learning}.
At the most basic approach, a prompt can take the form of a sentence ``a photo of a \texttt{[CLASS]}" where \texttt{[CLASS]} can be a category such as ``bird" or ``car". Formally, we can define the image encoder as \(f_{\phi}\) and text encoder as \(f_{\psi}\). Given an input image \(x\) and a set of \emph{static} prompts \(Y = \{y_1, y_2, ..., y_k\}\) where \(k\) is the number of categories, the image features \(z\) can be extracted such that \(z = f_{\phi}(x) \in \mathbb{R}^d\) and text features \(W = f_{\psi}(Y) \in \mathbb{R}^{d\times k}\). Lastly, the predicted category probabilities can then by computed by taking the softmax over the cosine similarities such that:

\begin{equation}\label{eq:cosine_sim}
p( y = i | x) = \frac{\exp(\cos(w_i, z) / \tau)}{\sum^k_{j=1} \exp(\cos(w_j, z) / \tau)}   
\end{equation}

where \(\tau\) is a fixed temperature hyperparameter, \(\cos(\cdot, \cdot)\) the cosine similarity, and $y$ the predicted prompt. By formulating the categories in such a way, CLIP can effectively be used as a versatile classification tool without task-specific training.

\paragraph{Few-Shot learning with CLIP}

In the few-shot learning scenario, the goal is to adapt the CLIP model to perform better on a target task using a limited number of labeled examples per class. To achieve this, we introduce a learnable component into the model, which can be fine-tuned on the few-shot training data. This learnable component can take various forms, such as learnable vectors added to the input prompts (like in CoOp and VPT) or learnable layers processing the visual and/or text embeddings, like in our proposed method.

\subsection{Adaptive Prompt Tuning}\label{sec:apt_method}
As shown in the preliminary experiments in \Cref{sec:experiments}, a static approach to prompt tuning is insufficient for datasets characterized by high intra-class variance in the image features.
To address these challenges, we propose an \emph{adaptive} prompting technique based on the cross-attention mechanism.
This allows for real-time adjustment of text prompts in response to relevant visual information in the test image at hand. 
\Cref{fig:architecture} illustrates the general architecture of the proposed model.

To address these challenges, we propose an \emph{adaptive} prompting technique based on the cross-attention mechanism.
This allows for real-time adjustment of text prompts in response to relevant visual information in the test image at hand. 
\Cref{fig:architecture} illustrates the general architecture of the proposed model. 

The architecture begins with a ViT, which processes an input image by dividing it into a sequence of patches that are flattened. These patches are then encoded into image features, with the output from the ViT including the \texttt{[CLS]} token, which captures the global representation of the image. \\

Parallel to the ViT processing of the image, the text encoder processes textual descriptions, typically prompts like ``A photo of a [CLASS].'' Formally, the text encoder \( f_{\psi} \) maps the input text \( Y \) to text features:

\begin{equation*}
W = f_{\psi}(Y) \in \mathbb{R}^{d\times k}
\end{equation*}

where \( d \) is the dimensionality of the feature space, and \(k\) the number of classes.\\

The core innovation of our model lies in the cross-attention module, where the encoded image features are used to refine the text features dynamically based on the visual content. 
The module integrates the image features $z$ extracted by the image encoder and the text features $W$ extracted by the text encoder.
Specifically, keys $k$ and values $v$ are created from $z$, while queries $q$ are created from $W$.
This mechanism allows the model to focus on relevant parts of both text and visual information, enabling dynamic adjustment of the text prompt based on visual information.

The cross-attention module additionally includes, as in the original implementation by \cite{vaswani2017attention}, layer normalization, a feed-forward layer---responsible for non-linearity within the module---skip connections, and dropout after the cross attention operation and the feed-forward layer.
The dropout modules, in addition to acting as a regularizer, enable us to perform UQ with MCD.

The output of the cross-attention module is a set of tuned text features $W^\prime$, which provide a refined representation optimized for downstream tasks.

\subsection{Comparable Methods}
\paragraph{CoOp} Context Optimization \cite{zhou2022conditional} is a method for few-shot learning in conjunction with CLIP. It utilizes the input tokens of the encoded text prompts. Specifically, it uses the tokens of the context prompt (e.g. ``A photo of...") before encoding and makes those learnable. During training this freezes the visual encoder, but adjusts the class embeddings to maximize the cosine similarity between the adjusted class embeddings and the training images of the respective class. However this requires to backpropagate through the full text encoder and limits the adjustability by the encoder itself.

\paragraph{VPT} Visual Prompt Tuning focuses on the visual encoder of CLIP. Instead of fine-tuning the weights of each transformer layer inside the encoder, it adds additional learnable tokens to the Transformer layer during training and inference.
\begin{equation*}\label{eq:clip-clf}
X = [P_1, \ldots, P_K, x_1, \ldots, x_N]
\end{equation*}
Where $X$ is the set of input token of each transformer layer, $x_n$ input tokens, which can be a patch embedding or an output token of the previous Transformer layer. $P$ is a set of learnable tokens, which is unique for each Transformer layer.

\subsection{Uncertainty Quantification with Monte-Carlo Dropout}

As explained in \Cref{sec:introduction}, UQ is essential for assessing model reliability and improving decision-making processes, especially in applications requiring high assurance in prediction accuracy. 

In particular, techniques like Monte Carlo (MC) Dropout have not been widely applied to CLIP or similar vision-language models. 



The presence of Dropout modules into APT allow us to implement MCD by simply avoiding to switch off the random dropout behavior at inference time.
As a consequence, the output of APT is not deterministic anymore.
We can use Monte-Carlo sampling to obtain an output probability distribution.
Given a sample size $M$, we can calculate an output mean for class $k$:

\begin{equation*}
\bar{p}_k = \frac{1}{M} \sum_{m=1}^{M} p_k^{(m)},
\end{equation*}

where $p_k^{(m)}$ indicates the output probability for class $k$ at sample $m$.
Then, the predicted class $\hat{y}$ is obtained as

\begin{equation*}
\hat{y} = \arg \max_k \bar{p}_k
\end{equation*}


We can estimate the uncertainty by means of entropy:

\begin{equation*}
S = - \sum_{k} \bar{p}_k \log \bar{p}_k.
\end{equation*}

Entropy is not the only way to quantify uncertainty; other methods, such as using the maximum predicted probability \(\max_k (\bar{p}_k)\), are also available. However, we select entropy because it provides a comprehensive measure of uncertainty across all possible classes and not only the predicted class.
To convert entropy into confidence, we can normalize entropy such that it lies in the $[0,1]$ interval, then subtract 1 to it:

\begin{equation*}
    \text{conf} = 1-\frac{S}{\max_S},
\end{equation*}

where $\max_S$ indicates the maximum value of entropy over a given sample of data.



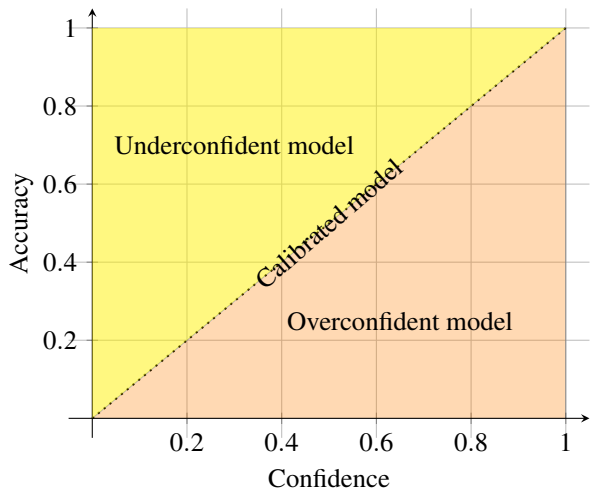
\begin{figure}
    \centering
    \input{figs/diagrams/reliabliity}
    \caption{Reliability plots help with qualitative evaluation of uncertainty.
    Points below the dotted diagonal line indicate overconfident behaviors (where confidence is higher than accuracy); above the diagonal line instead is the area of underconfidence.
    For calibrated models, the points will roughly lie around the diagonal line.
    }
    \label{fig:example_reliability}
\end{figure}

The uncertainty estimates are determined to be optimal when there is an equality between accuracy and confidence, so prediction confidence---which is known at inference time---can be used as a proxy for accuracy---which is unknown in absence of labels---thus allowing to discard potentially inaccurate predictions.
The assessment of the uncertainty estimates happens qualitatively by means of reliability plots, which plot confidence vs.~accuracy.
For a given validation dataset, the confidence is split in bins and the mean per-bin accuracy is computed.
The resulting plot indicates whether the model is overconfident (confidence $>$ accuracy), underconfident (confidence $<$ accuracy) or calibrated (confidence $\approx$ accuracy), as depicted in \Cref{fig:example_reliability}.

A quantitative evaluation of uncertainty is possible with the Expected Calibration Error (ECE), which builds upon the setup of reliability plots.
Given bins $B_1,\dots,B_P$, ECE computes the absolute value of the per-bin difference between accuracy and confidence, then sums these values over the $P$ bins:

\begin{equation*}
\text{ECE} = \sum_{p=1}^{M} \frac{|B_p|}{N_p} \left| \text{acc}(B_p) - \text{conf}(B_p) \right|,
\end{equation*}

where $\text{acc}(B_p)$ and $\text{conf}(B_p)$ indicate, respectively, the accuracy and the confidence of the data points in bin $B_p$, while $N_p$ indicates the number of data points belonging to bin $B_p$.
An ECE approaching 0 signals a case of perfectly calibrated model, while a high ECE points at under-- or over-confidence, or a mix of the two.

In addition to reliability plots and ECE, we produce another qualitative analysis by means of Confidence vs.~Uncertainty plots.
Given a dataset, we plot incorrect and correct predictions.
From a visual perspective, we would expect to see most of the incorrect predictions in the bottom-right portion of the plot (i.e., where uncertainty is high and confidence is low), while correct predictions should concentrate in the upper-left area, potentially with minimal overlap between the two categories.

\paragraph{Out-of-Distribution detection}

As an additional assessment on the uncertainty estimates, we perform an Out-of-Distribution (OOD) detection analysis on our model.
Several works \cite{nguyen2015deep,valdenegro2021find} have shown that Deep Neural Networks tend to produce extremely overconfident predictions on OOD---sometimes even random---data.
Nevertheless, OOD data often occurs in real-life scenarios: the presence of outliers, scenarios not accounted for in the designing phase of a model, or distribution shifts are all phenomena that contribute to this factor.
In the presence of OOD data, we expect our model to produce low-confidence predictions.
Conversely, in the case of in-distribution data, we expect the model to output highly-confident predictions.
The aim is to possibly identify ways to tell apart OOD and in-distribution data by means of the predictive uncertainty.
We qualitatively perform the OOD detection analysis by means of Confidence vs.~Entropy plots, where we expect to see the distribution concentrated in the bottom-right part, with correct and incorrect predictions largely overlapping, due to the model not having any notion of what would constitute ``correctness'' in the case of an OOD data.

\subsection{Datasets}

To evaluate the performance of the proposed vision-guided prompting approach, we make use of popular datasets in the field of fine-grained few-shot classification, each chosen for its specific characteristics and the unique challenges it presents. 


The Caltech-UCSD Birds \textbf{CUBIRDS} dataset \cite{WahCUB_200_2011} contains \num{11788} images across \num{200} categories of birds.

The \textbf{Oxford Flowers} dataset \cite{nilsback2008automated} features \num{8189} images of flowers split into \num{103} classes, each having from 40 to 258 samples.

The \textbf{FGVC Aircrafts} dataset \cite{maji13fine-grained} includes \num{10200} pictures of aircraft divided into 102 categories, each holding 102 images.

We make use of these three datasets for assessing the models capabilities in fine-grained classification tasks.

Additionally, we use the \textbf{Caltech101} dataset for assessing the model OOD detection capabilities.
This dataset contains \num{9146} images from 101 heterogeneous categories, featuring between 45 to 800 samples each.

\subsection{Implementation Details}

In our implementation, we made use of CLIP with a ViT-B/16 image encoder. 
Both text and image encoder used multi-head attention layers with 8 heads.
As for APT specifically, we applied dropout with a rate of 20\%.
While this value may technically be considered low for standard dropout, for MCD we followed the indications by \cite{seoh2020qualitative} to keep a lower dropout rate.
We followed the standard image preprocessing steps used by CLIP, as well as data augmentation techniques from CoOp, and VPT, these being random resized cropping and flipping.
Also, similarly to the CoOp and VPT set ups, we trained the models over 50 epochs with 1 sample per class, 100 epochs for 2 to 4 samples per class, and 150 epochs for 8 to 16 samples per class.
We made use of the SGD optimizer with a learning rate of 0.001 and a cosine decay learning rate scheduler.
For each combination of model (APT, CoOp, VPT), dataset, and samples per class, we repeated the training 3 times and report the average performance attained.
We performed all experiments on one NVIDIA A100 GPU.

\section{Results}

\subsection{Few-Shot Learning}
\begin{figure*}[!h]
    \centering
    \begin{tikzpicture}[scale=1]
        \node at (11.475,0) {
            \includegraphics[width=0.25\linewidth, trim=0cm 0cm 1cm 1cm, clip]{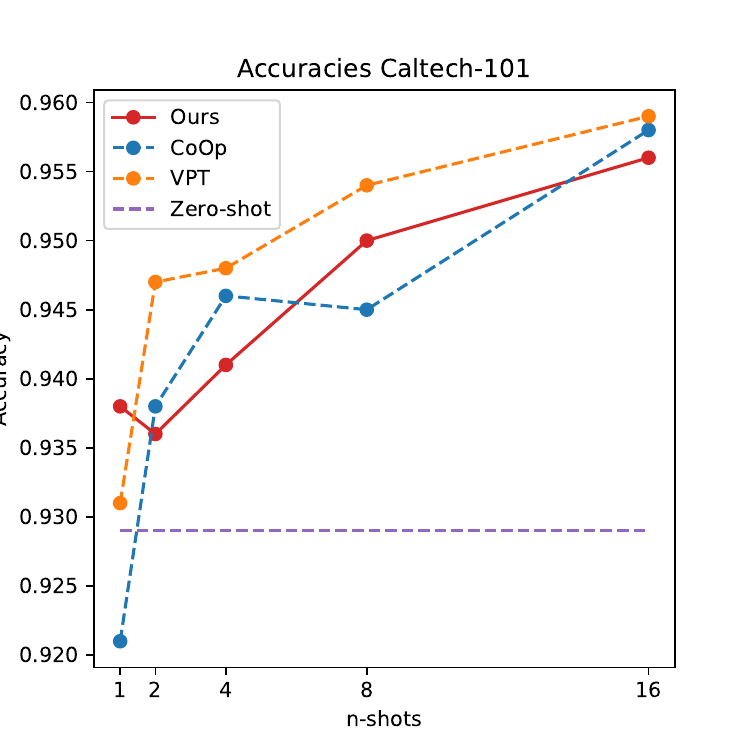}
        };

        \node at (7.6,0) {
            \includegraphics[width=0.25\linewidth, trim=0cm 0cm 1cm 1cm, clip]{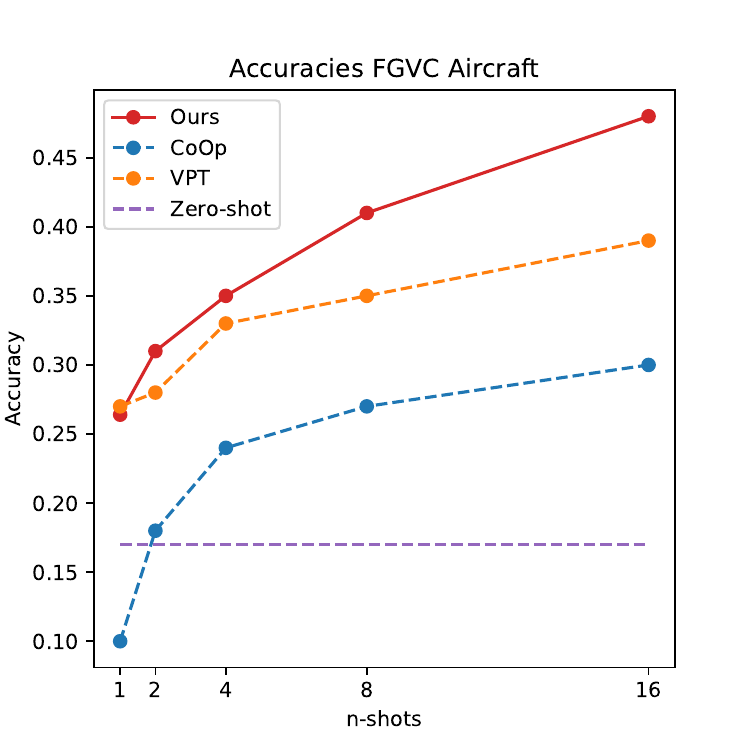}
        };

        \node at (3.8,0) {
            \includegraphics[width=0.25\linewidth, trim=0cm 0cm 1cm 1cm, clip]{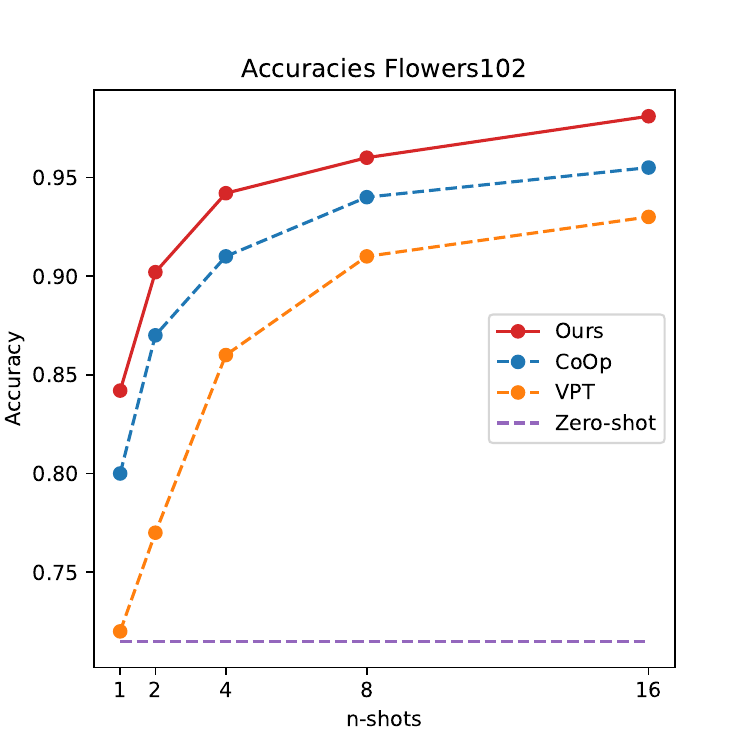}
        };
        
        \node at (0,0) {
            \includegraphics[width=0.25\linewidth, trim=0cm 0cm 1cm 1cm, clip]{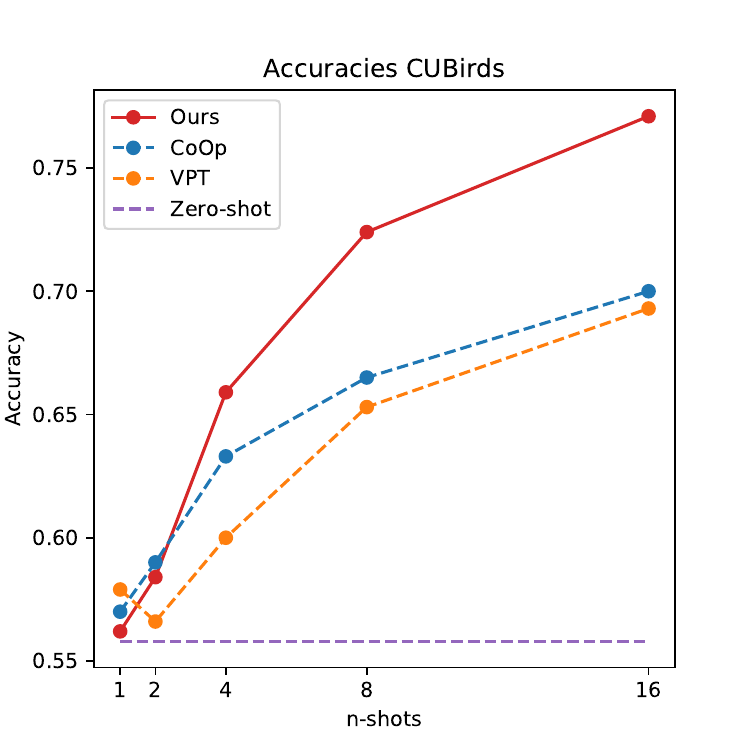}
        };

    \end{tikzpicture}
    \vspace{-6pt}
    \caption{Results of the few-shot learning set up. Our approach (red) is compared to the baseline CLIP results (purple), CoOp (blue), and VPT (yellow). Results are from the average scores of 3 models, where the training images are sampled with different sampling seeds.}
    \label{fig:plots-fsl}
\end{figure*}

In \Cref{fig:plots-fsl} we depict the results in terms of accuracy for APT, CoOp, and VPT.
As a baseline, we additionally report the performance of CLIP used as a zero-shot classifier.

Across the FGVC Aircraft dataset, which exhibits high intra-class variance, our cross-attention model outperformed others from 2 to 16 shots. Starting with 27\% accuracy at one shot and rising to 47\% at 16 shots, it showed a significant improvement over the Zero-shot baseline (17\%). This demonstrates the model's  ability to successfully leverage additional examples. CoOp and VPT also improved, but not as markedly, highlighting the strength of our approach in handling complex variations within the dataset. In contrast, the Oxford Flowers dataset, with distinct inter-class features, showed strong performance across models. Our model reached 84\% accuracy with one shot, improving to 97\% at 16 shots. CoOp closely followed, while VPT lagged behind. The baseline of 72\% further highlights the significant enhancement brought by few-shot learning. Lastly, the CUBirds dataset, with both high inter- and intra-class variance, revealed our model’s robustness. Starting at 56\% accuracy with one shot and reaching 77\% at 16 shots, the model outpaced CoOp and matched the Zero-shot baseline at one shot, showing minimal learning without additional examples.

For the Caltech101 dataset, our model did not show significant improvement over the other fine-tuning approaches, such as CoOp and VPT, the exception being in the 1-shot setting. The accuracy improvements with increasing shots were relatively modest, indicating that the dynamic nature of the cross-attention model might not have provided a substantial advantage in this more general classification setting. Specifically, APT showed a gradual increase in accuracy from 93.8\% at one shot to approximately 95.8\% at 16 shots. This performance was comparable to that of CoOp and VPT-deep, which also exhibited similar improvement trends with increased shots. The modest improvement on the Caltech101 dataset can be explained by its more general nature compared to fine-grained datasets like FGVC Aircraft or CUBirds. Since the classes are more distinct and easier to differentiate, the complexity of dynamic prompt adjustment may not provide the same benefit as in datasets with subtle inter-class differences. Static prompts and visual tokens may suffice for high performance here. Additionally, Caltech101’s varied objects and backgrounds can introduce noise, potentially causing adaptive prompt tuning to focus on irrelevant features and affecting performance.

Overall, our cross-attention model performs robustly across varying degrees of class variances, showing adaptability and improved accuracy with increased shots across all datasets. However, its performance on the Caltech101 dataset, which features a wider variety of objects and backgrounds, highlights a critical aspect of this approach. The adaptive prompt tuning can be influenced by background noise, leading to less optimal tuning in general classification settings. This highlights the importance of understanding and leveraging both intra-class and inter-class variances, as well as the dataset's contextual complexity, to optimize model performance. These results suggest that while adaptive prompt tuning offers significant advantages in fine-grained classification tasks, its benefits must be carefully weighed against potential limitations in more general or cluttered environments.

\paragraph{Generalization Setup}

\Cref{tab:basetonew} summarizes the performance of different models on the base and new classes across the FGVC Aircraft, Oxford Flowers, and CUBirds datasets. The metrics reported include the accuracy on the base classes (Base), the accuracy on the new classes (New), and the harmonic mean (F1-Mean).

\begin{table*}[!h]
    \caption{Results (accuracy) of the base to new class setup. The results are produced from the average of 3 sampling seeds of the training set, where 16 samples were taken per class.}
    \centering
    \begin{tabular}{|l|c|c|c| c|c|c| c|c|c|}
        \hline
& \multicolumn{3}{l|}{FGVC Aircraft} & \multicolumn{3}{l|}{Oxford Flowers} & \multicolumn{3}{l|}{CUBirds}\\
        \hline
        Model & Base & New & F1 & Base & New & F1 & Base & New & F1\\
        \hline
         CLIP & 27.19 & \textbf{36.29} & 31.09 & 72.08 & \textbf{77.80} & 74.83 & 65.18 & \textbf{52.34} & \textbf{58.06}\\
         CoOp & 40.44 & 22.30 & 28.75 & 97.60 & 59.67 & 74.06 & 81.51 & 34.63 & 48.60 \\
         CoCoOp & 33.41 & 23.71 & 27.74 & 94.87 & 71.75 & 81.71 & 71.97 & 08.04 & 14.40 \\
         APT (ours) & \textbf{43.74} & 31.26 & \textbf{36.46} & \textbf{98.64} & 71.98 & \textbf{83.23} & \textbf{83.02} & 43.42 & 57.02\\
         \hline
    \end{tabular}
    \vspace*{-1.25em}
    \label{tab:basetonew}
\end{table*}

For the FGVC Aircraft dataset, the cross attention model demonstrated a notable improvement over other models. Specifically, it achieved an F1-Mean score of 36.46, outperforming both CoOp and CoCoOp models, which scored 28.75 and 27.74, respectively. This improvement highlights the model's ability to generalize well to new classes, which is critical given the high intra-class variance within this dataset. The cross attention model's dynamic adaptation through feedback from image features likely contributed to its superior performance, allowing it to capture subtle distinctions between different aircraft types effectively. \\

In the Oxford Flowers dataset, characterized by distinct and less variable class features, the cross attention model again led the performance metrics. It achieved an F1-Mean score of 83.23, significantly higher than CoOp's 74.06 and CoCoOp's 81.71. The ability to generalize to new classes while maintaining high accuracy underscores the model's robustness. The results suggest that incorporating image features into the prompt, as done in CoCoOp, provides an advantage, but the cross attention mechanism further enhances this by dynamically adjusting to new visual inputs, leading to superior generalization. \\

For the CUBirds dataset, the cross attention model showed strong generalization capabilities with an F1-Mean score of 57.02. This was competitive with the baseline CLIP model, which had an F1-Mean of 58.06, but it outperformed both CoOp (48.60) and CoCoOp (14.40) models. Notably, the cross attention model excelled in the base class performance, achieving the highest accuracy among all models. However, the generalization to new classes was less pronounced compared to FGVC Aircraft and Oxford Flowers. This indicates that while the model adapts well within known classes, the complexity and similarity of bird species present a more significant challenge for unseen classes.

\subsection{Uncertainty Quantification}

\begin{figure}[!h]
    \centering
    \includegraphics[width=\linewidth,trim=1cm 1cm 0 1cm]{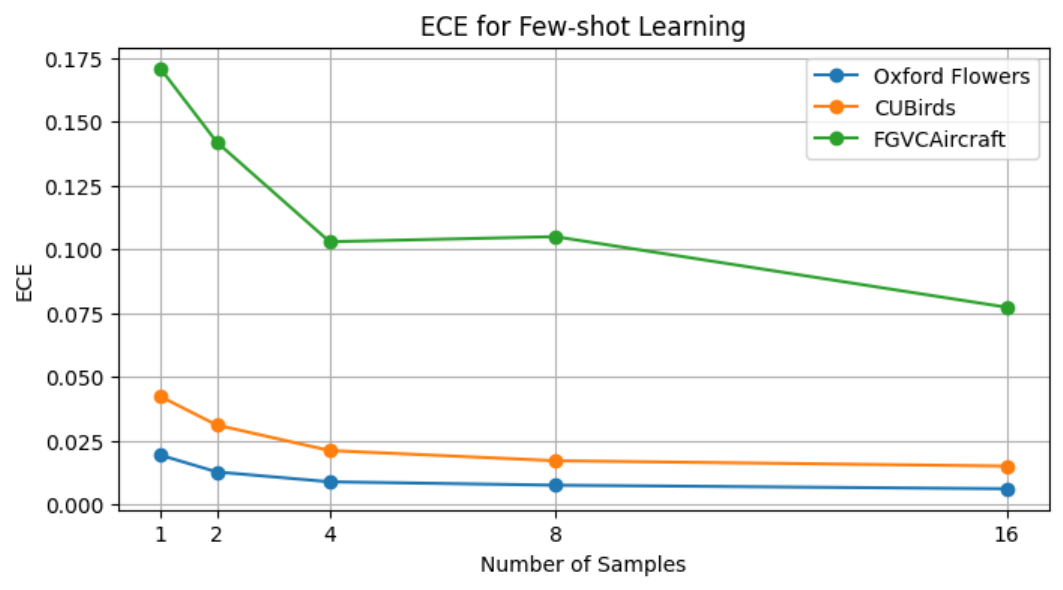}
    \vspace{0.0cm}
    \caption{Expected Calibration Error (ECE) across number of training samples. A lower ECE indicates a better calibration. It can be observed that as the number of samples is increased, the ECE decreases.}
    \label{fig:ece-fsl}
\end{figure}

\paragraph{Expected Calibration Error} 
The ECE for few-shot learning across different datasets as function of the number of samples provides an insight into the model's reliability of its confidence estimates with limited training data (\Cref{fig:ece-fsl}). \\ 

For the Oxford Flowers and the CUBirds datasets, the ECE shows a consistent decrease as the number of samples increases, stabilizing at values around 0.01 and 0.02. This trend suggests that the model remains well-calibrated as more samples are provided.
The more gradual improvement in ECE for the FGVC Aircraft dataset indicates that it poses a more substantial challenge for calibration; nonetheless the model demonstrates improvement with additional samples, showing its potential for better calibration with more extensive training. \\


\begin{figure*}[!h]
    \centering
    \begin{subfigure}[b]{0.8\textwidth}
        \centering
        \includegraphics[width=\textwidth]{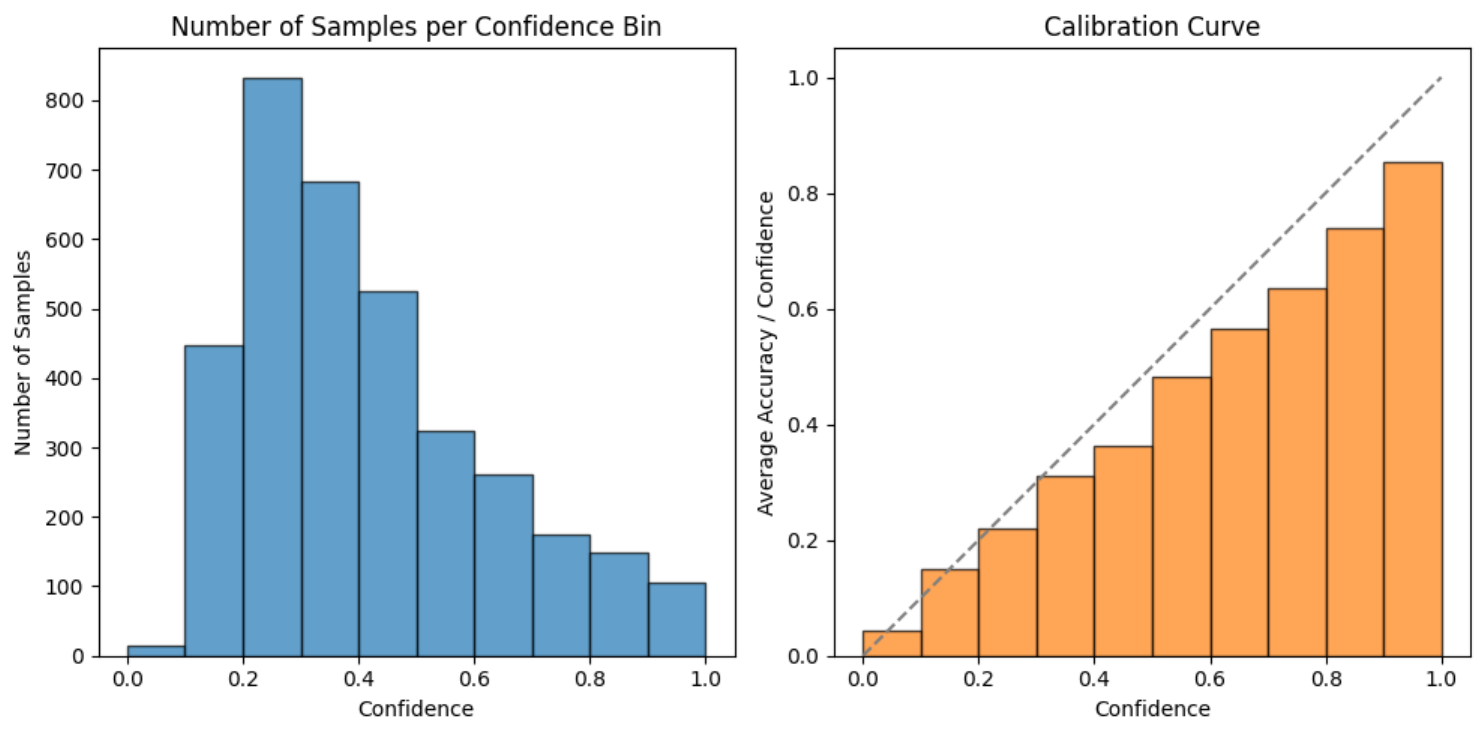}
        \caption{Calibration for the CUBirds dataset.}
        \label{fig:b}
    \end{subfigure}
    \vskip\baselineskip
    \begin{subfigure}[b]{0.8\textwidth}
        \centering
        \includegraphics[width=\textwidth]{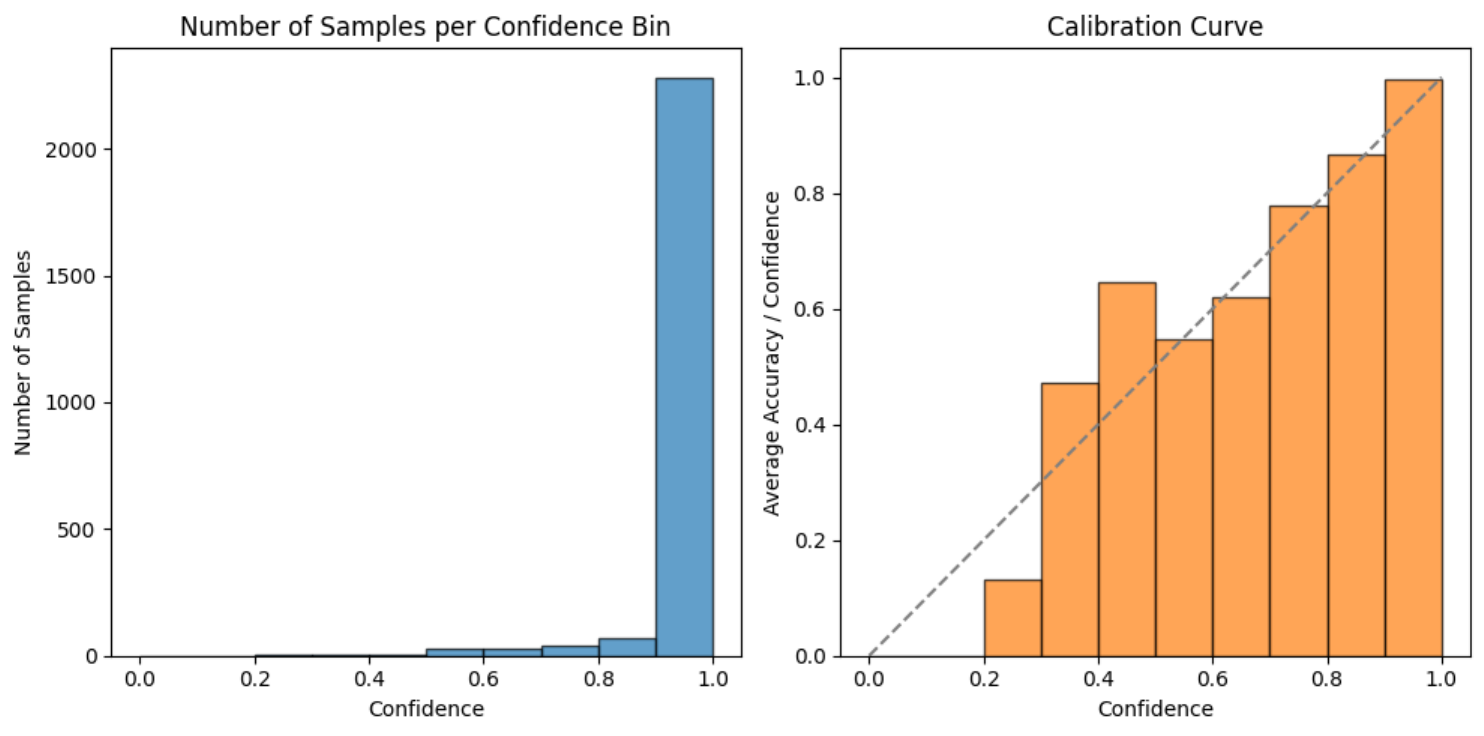}
        \caption{Calibration for the Oxford Flowers dataset.}
        \label{fig:d}
    \end{subfigure}
    \vskip\baselineskip
    \begin{subfigure}[b]{0.8\textwidth}
        \centering
        \includegraphics[width=\textwidth]{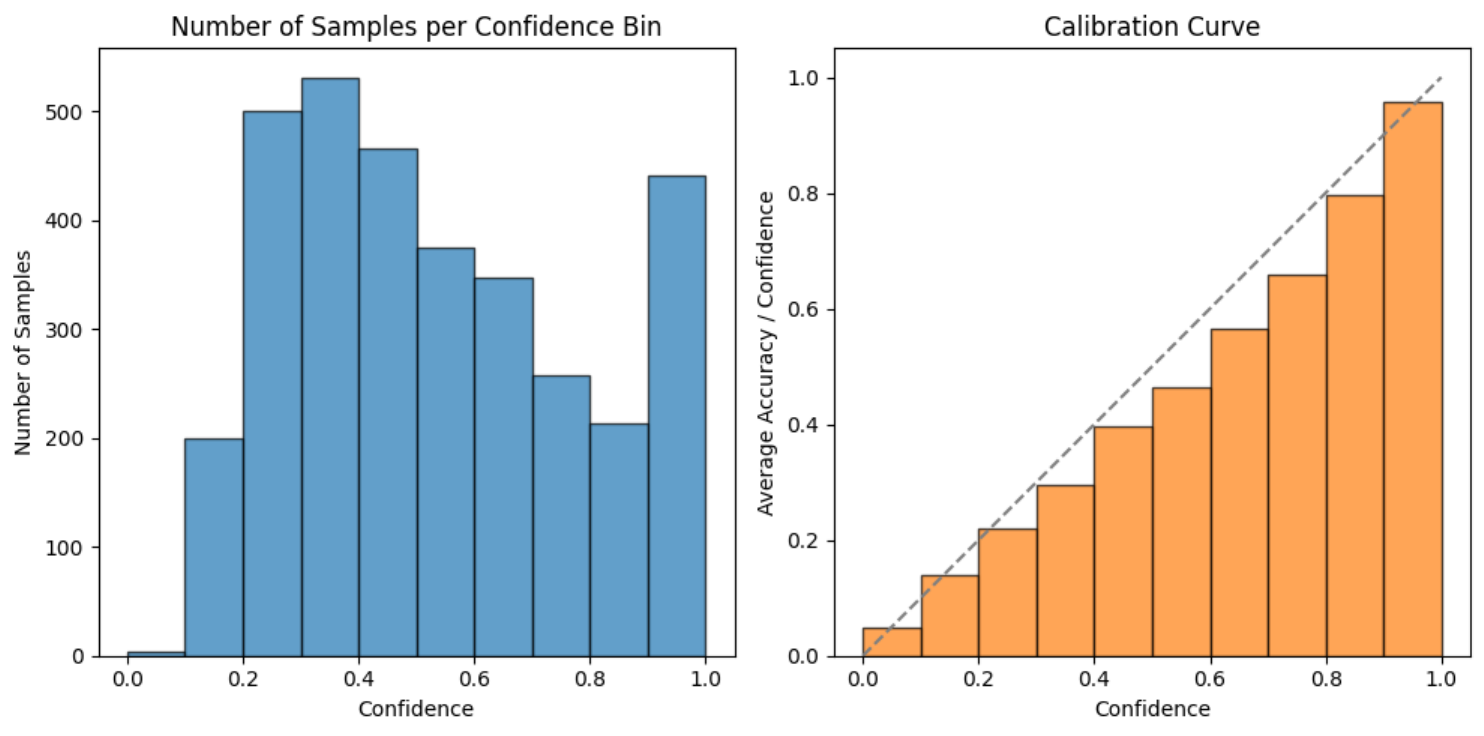}
        \caption{Calibration for the FGVC Aircraft dataset.}
        \label{fig:f}
    \end{subfigure}
    \caption{Calibration plots for different datasets showing the model's performance in predicting correct class probabilities. Each plot illustrates the relationship between the predicted probability and the actual accuracy, helping to assess the reliability of the model's probabilistic predictions.}
    \label{fig:ece_barplot}
\end{figure*}

\Cref{fig:ece_barplot} provides a more in depth visualization of the calibrations, which reveal how well the model's confidence aligns with its actual accuracy. These plots illustrate the accuracy versus confidence for the CUBirds, Oxford Flowers, and FGVC Aircraft datasets. Each dataset has been trained with a dropout rate of 0.2 and with 16 samples per class. Each plot compares the model accuracy to a perfectly calibrated model (black dashed line), providing insights into how well the model predictions match actual outcomes. \\

From these charts, we can see how the model is slightly overconfident on CUBirds and FVGC Aircraft, while we get a mixed profile for Oxford Flowers.
For CUBirds, there is a noticeable drop in accuracy for the high-confidence bins (0.9 to 1.0), with an average accuracy of about 15 point percentages lower than confidence.
On the flowers dataset, the model seems to be perfectly calibrated on the 0.7--1.0 bins, while it showcases under-- and overconfidence in the earlier bins.
This behavior is possibly due to the high accuracy of the model, which leaves few data points in the lower half of the confidence (and accuracy) spectrum.
The FVGC Aircraft has a lower overall accuracy, which cause the ECE to give more importance in the middle of the confidence spectrum, where most data points are located.
The chart shows how the model seems particularly overconfident in the 0.5--0.7 confidence range.



\paragraph{Confidence vs Uncertainty}

\begin{figure*}[t]
    \centering

    \begin{tikzpicture}
        \node at (6,0) {\includegraphics[width=.4\textwidth]{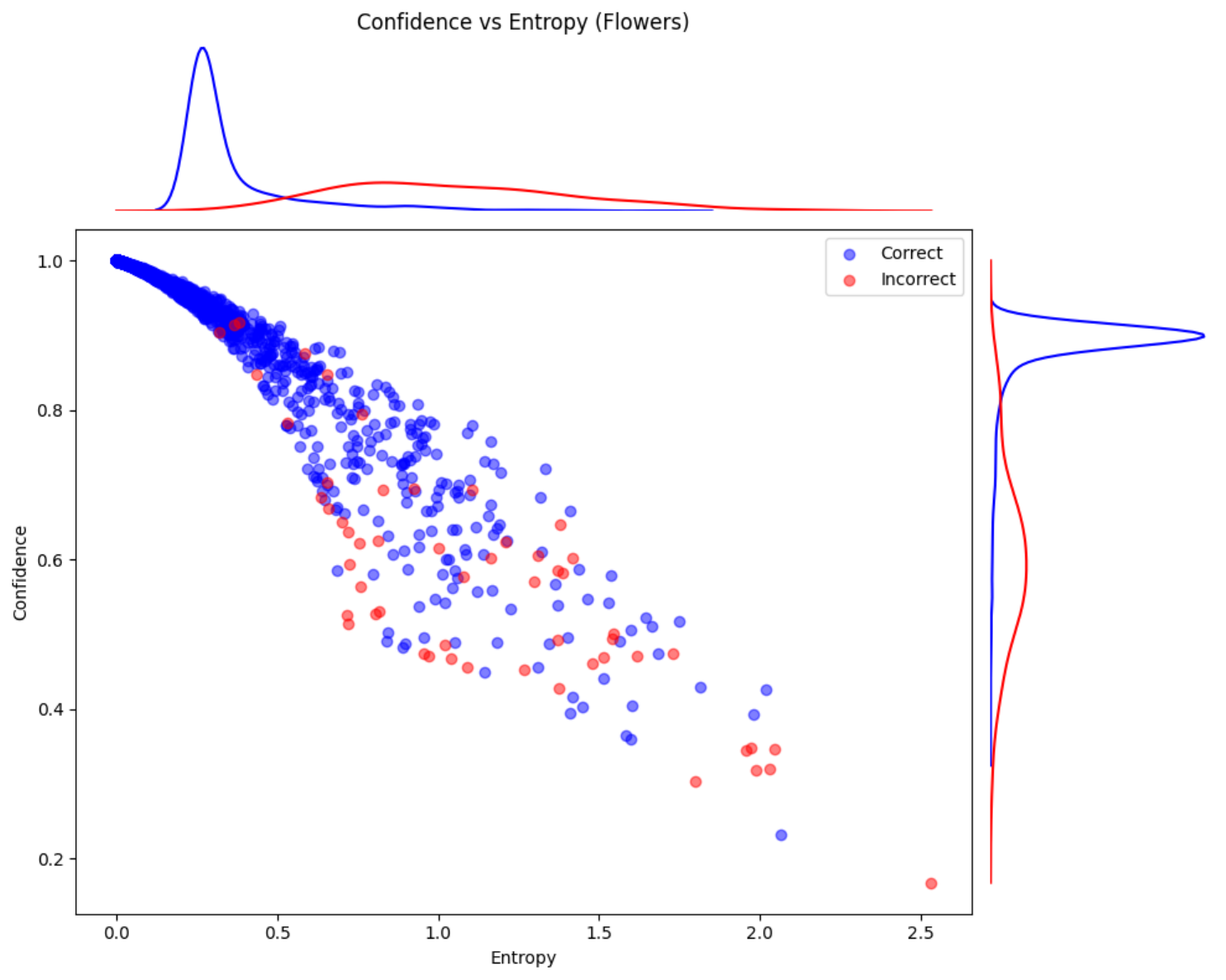}};
        \node at (0,0) {\includegraphics[width=.4\textwidth]{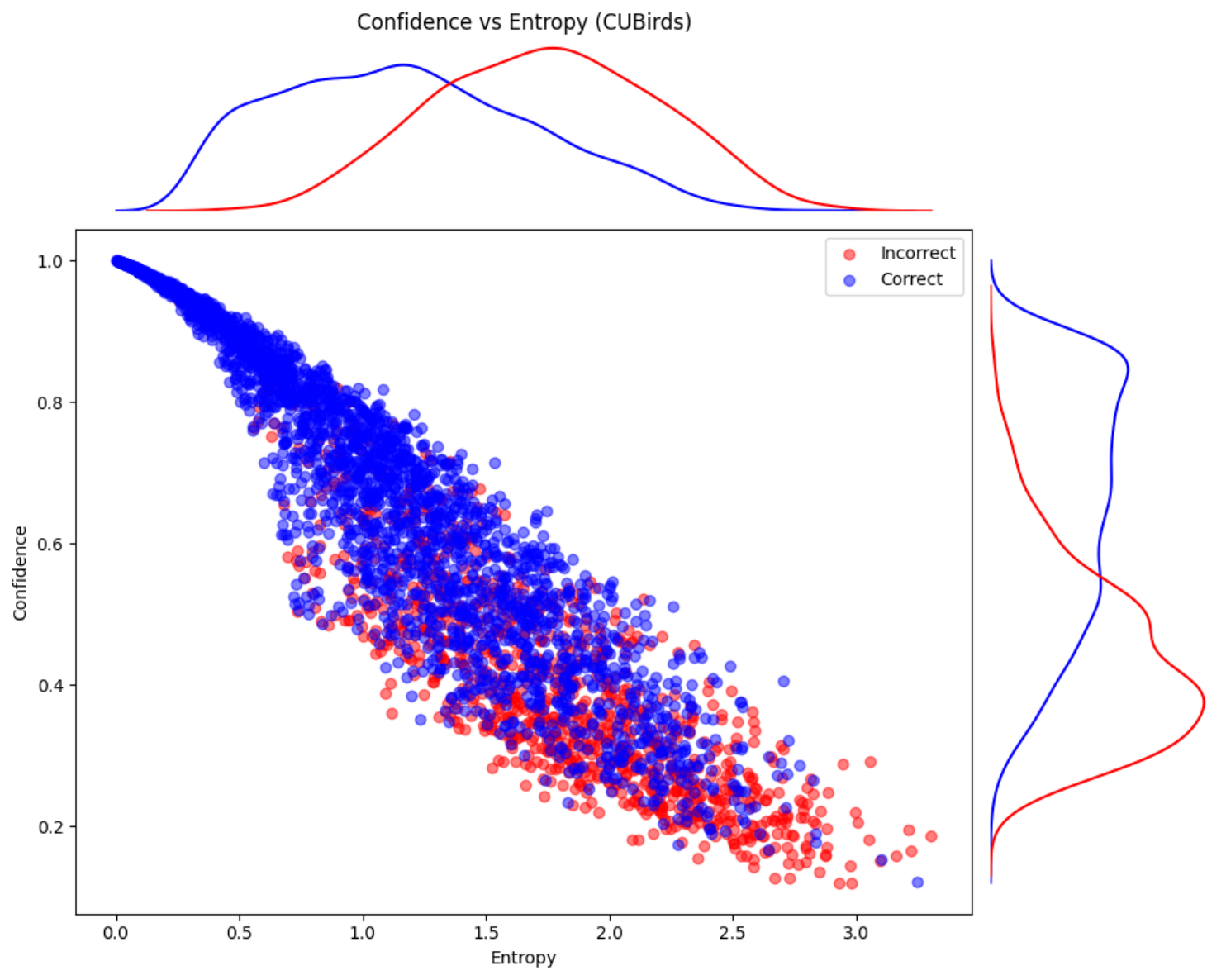}};
        \node at (6,-5.15) {\includegraphics[width=.4\textwidth]{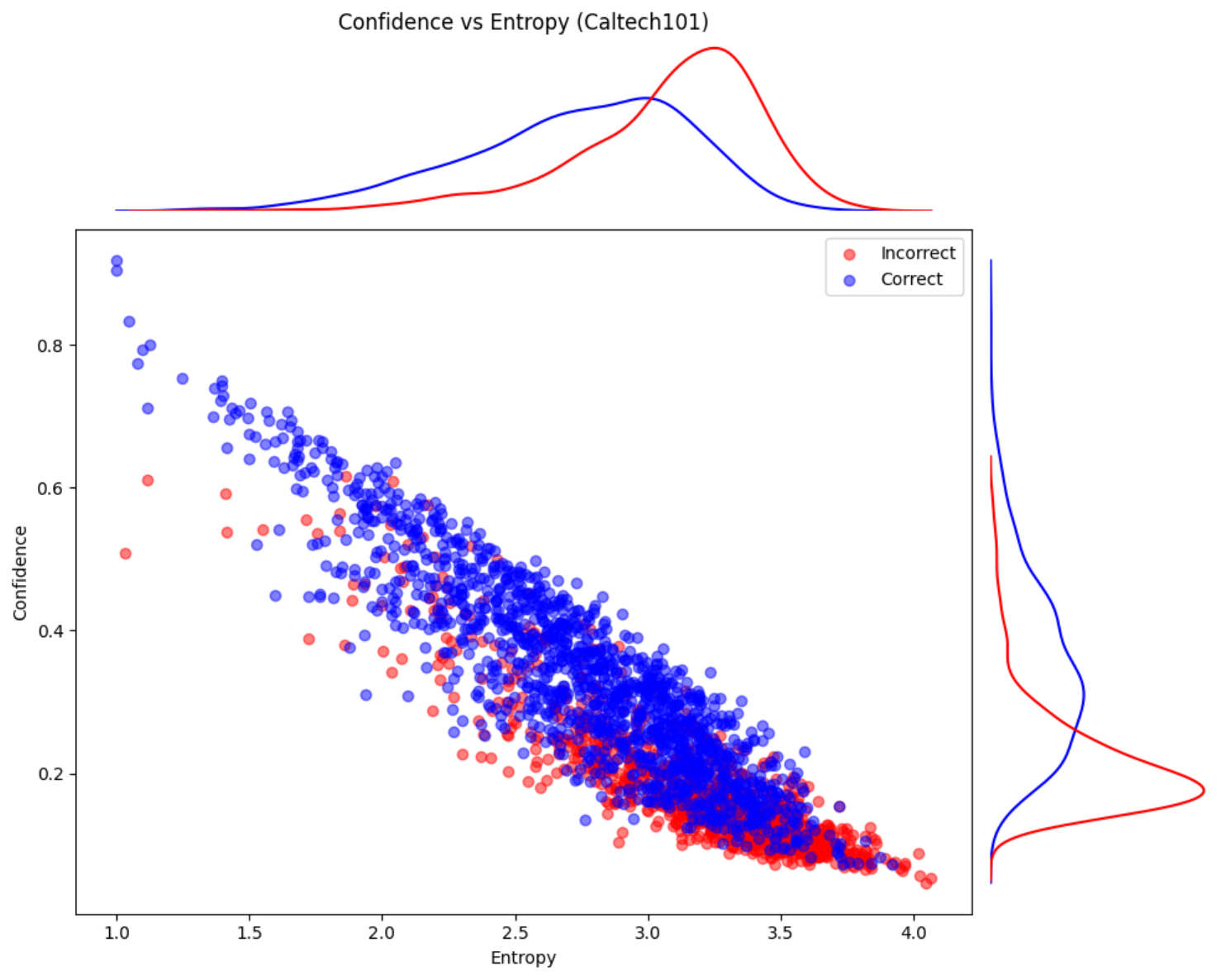}};
        \node at (0,-5.15) {\includegraphics[width=.4\textwidth]{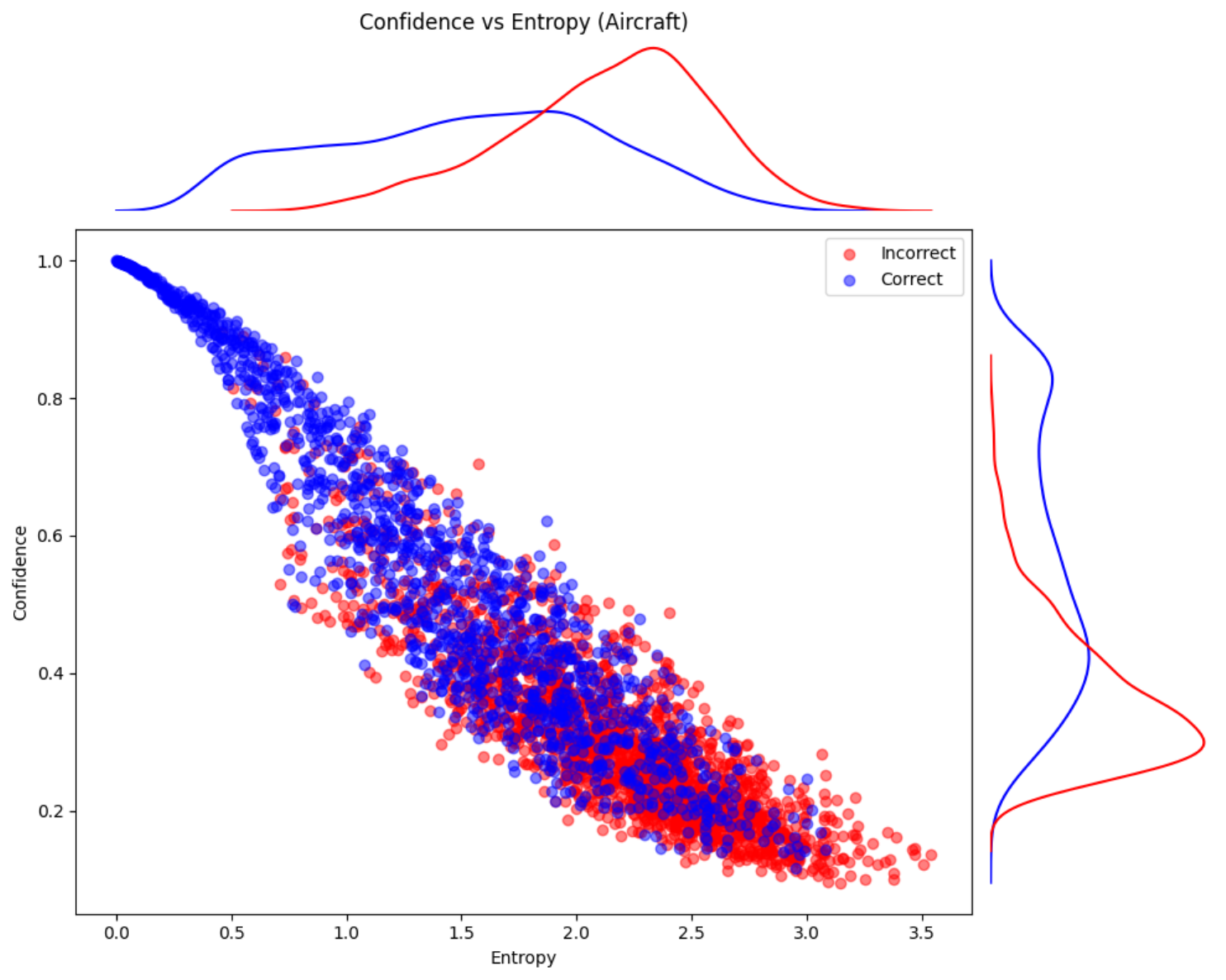}};
    \end{tikzpicture}
    
    \caption{Confidence vs. Uncertainty plots for the Oxford Flowers, FGVC Aircraft, and CUBirds datasets show an inverse relationship between prediction confidence and uncertainty. Correct predictions (blue) cluster at higher confidence and lower uncertainty, while incorrect predictions (red) are more dispersed. The KDE (Kernel Density Estimation) plots provide a visual representation of the density of predictions along the confidence and uncertainty axes, highlighting areas where predictions are most concentrated. This additional layer of information helps to identify the distribution of both correct and incorrect predictions across the confidence and uncertainty spectrum.}
    \label{fig:uncertainty}
\end{figure*}

The confidence vs uncertainty plots for the Oxford Flowers, FGVC Aircraft, and CUBirds datasets (\Cref{fig:uncertainty}) illustrate the relationship between the model's confidence in its predictions and the associated uncertainty. These plots help us understand how well the model can differentiate between certain and uncertain predictions, and how this differentiation impacts the accuracy of the predictions. The plots show the confidence and uncertainty of models trained with 16 samples per class. \\

Predictions with high confidence generally exhibit low uncertainty, while low-confidence predictions tend to have higher uncertainty, aligning with the model's ECE score and indicating accurate self-assessment. The inverse relationship between confidence and uncertainty is consistent across all datasets, with high-confidence predictions often correct and low-confidence ones more uncertain. The KDE density plots support this, showing that correct predictions cluster in high-confidence, low-uncertainty regions, while incorrect predictions are more dispersed with higher uncertainty. Notably, the FGVC Aircraft and CUBirds datasets show similar uncertainty spreads, despite CUBirds’ higher accuracy, possibly due to its larger number of classes, which increases entropy and uncertainty. \\

The ``cone shape" observed in the uncertainty vs. confidence plots reflects the model's ability to distinguish between easy and difficult examples. For easy cases, high confidence and low uncertainty align well with correct predictions, while more challenging examples maintain low uncertainty despite lower confidence, indicating the model’s awareness of difficulty. As uncertainty increases, confidence narrows uniformly, reflecting the model's caution in uncertain situations. In OOD settings, such as with the Caltech101 dataset, uncertainty remains consistently high, and predictions cluster around low confidence and high uncertainty, indicating the model's struggle to differentiate between correct and incorrect predictions in unfamiliar data, as expected for an OOD scenario.

\paragraph{Qualitative Analysis} 

\begin{figure*}
    \centering
    \begin{tikzpicture}[scale=.9, every node/.style={scale=0.9}]
        \node at (0,0) {\includegraphics[width=0.3\textwidth, trim=0.75cm 0.7cm 0cm .7cm, clip]{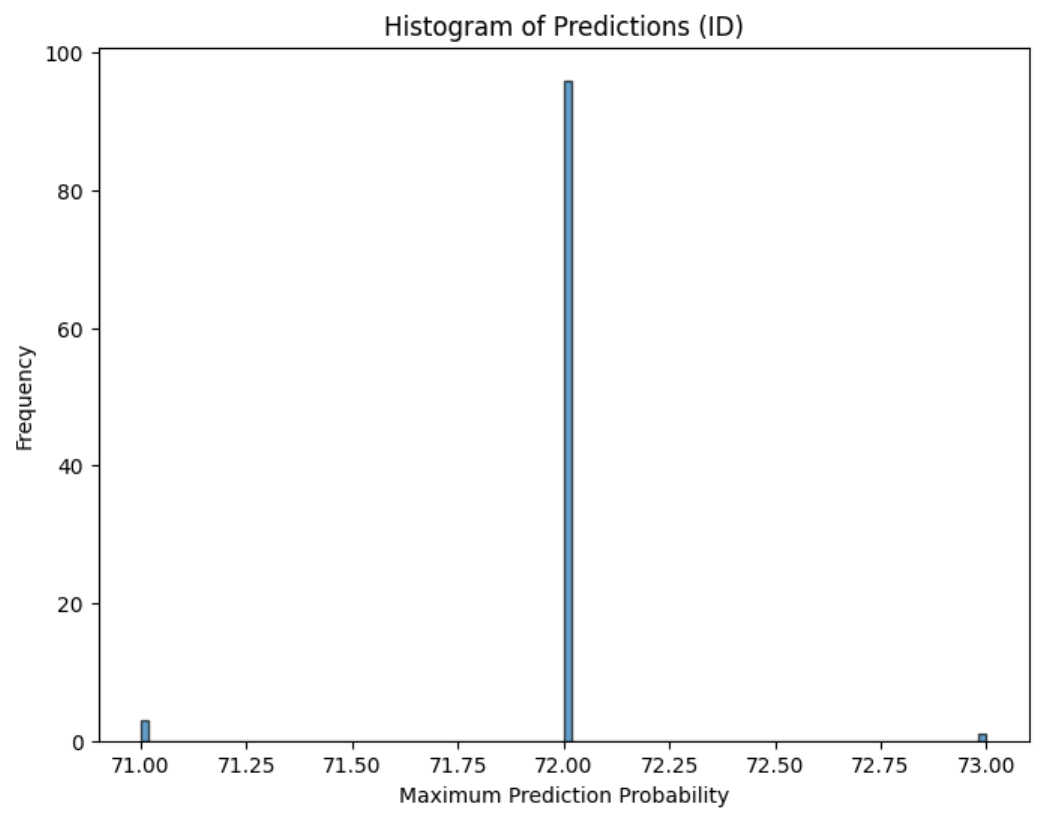}};
        \node at (5,0) {\includegraphics[width=0.3\textwidth, trim=0.75cm 0.7cm 0cm .7cm, clip]{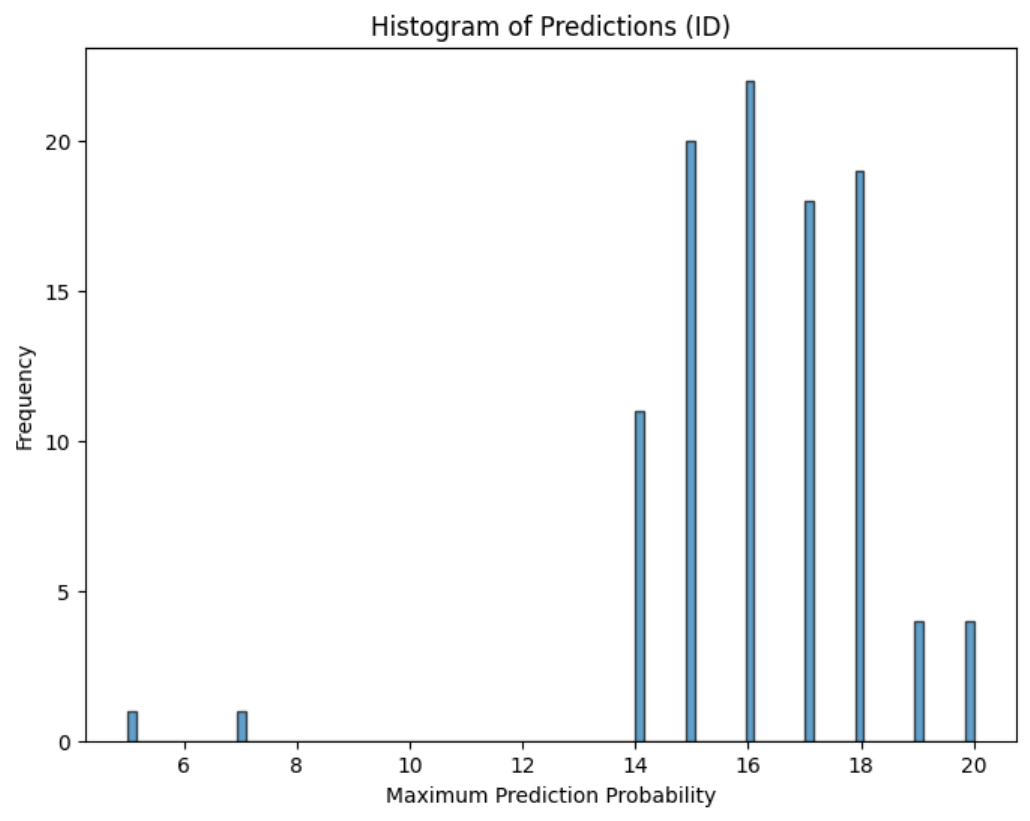}};
        \node at (10,0) {\includegraphics[width=0.3\textwidth, trim=0.75cm 0.7cm 0cm .7cm, clip]{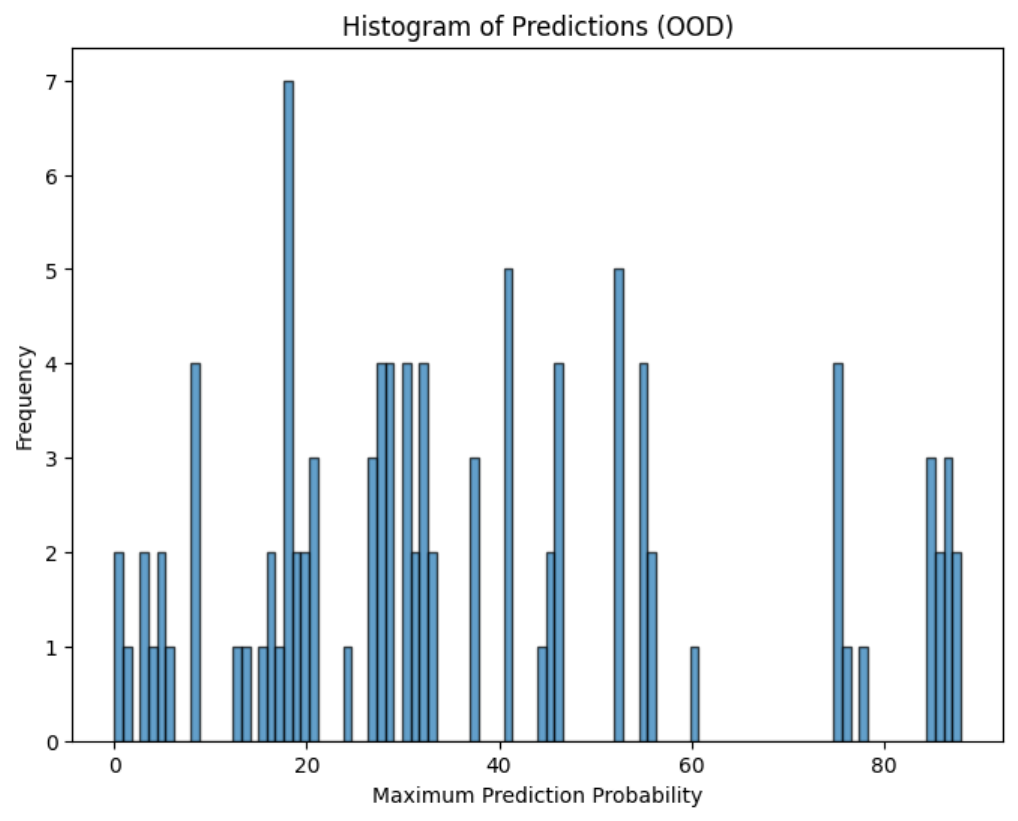}};

        \node at (5,-2) {\footnotesize Maximum Prediction Probability};
        \node[rotate=90] at (-2.5, 0) {\footnotesize Frequency};

        \node at (0,2) {\footnotesize in distribution};
        \node at (5,2) {\footnotesize in distribution};
        \node at (10,2) {\footnotesize OOD};

        \node at (0.25,-3.65) {\includegraphics[height=0.125\textheight]{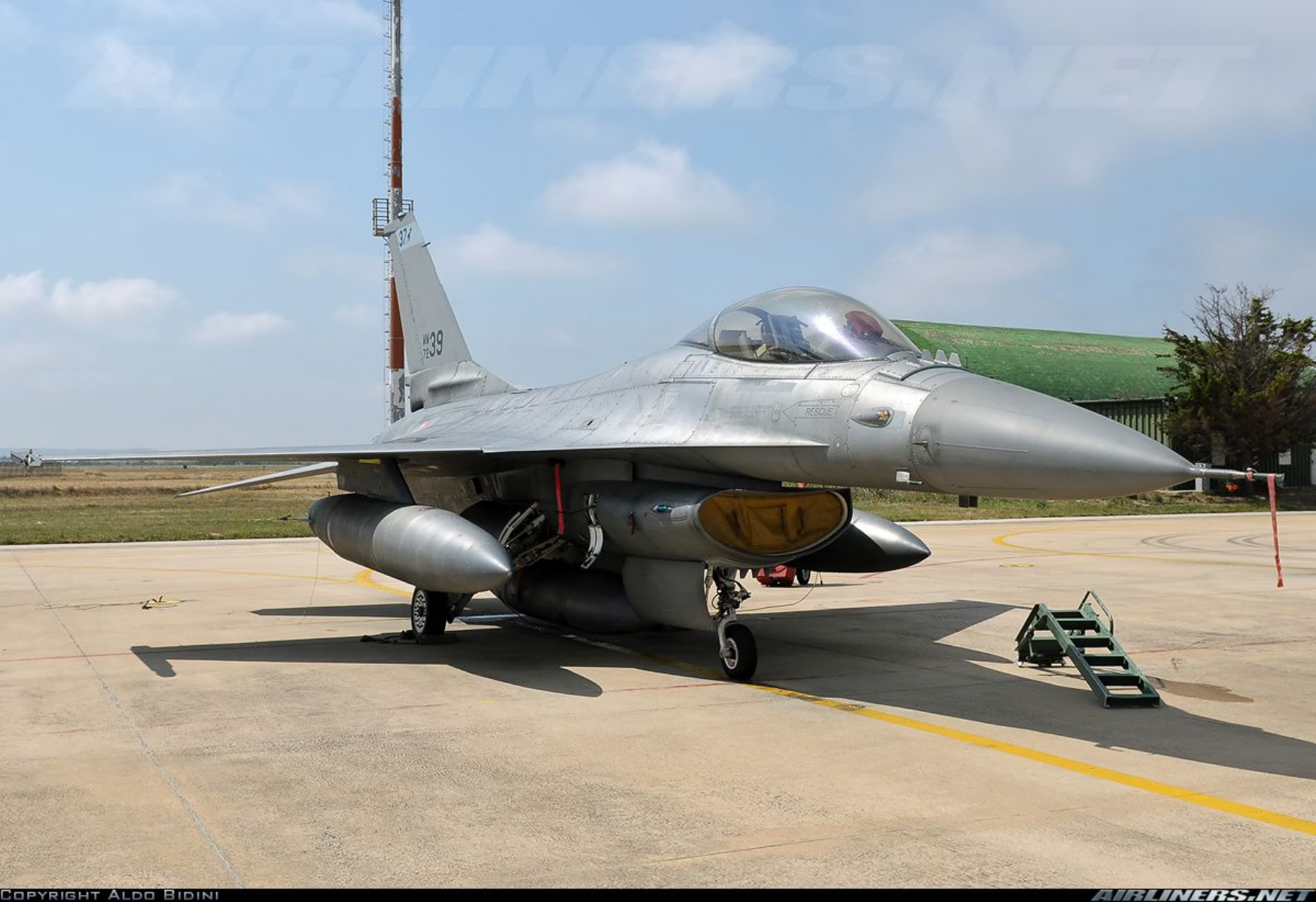}};
        \node at (5.2,-3.65) {\includegraphics[height=0.125\textheight]{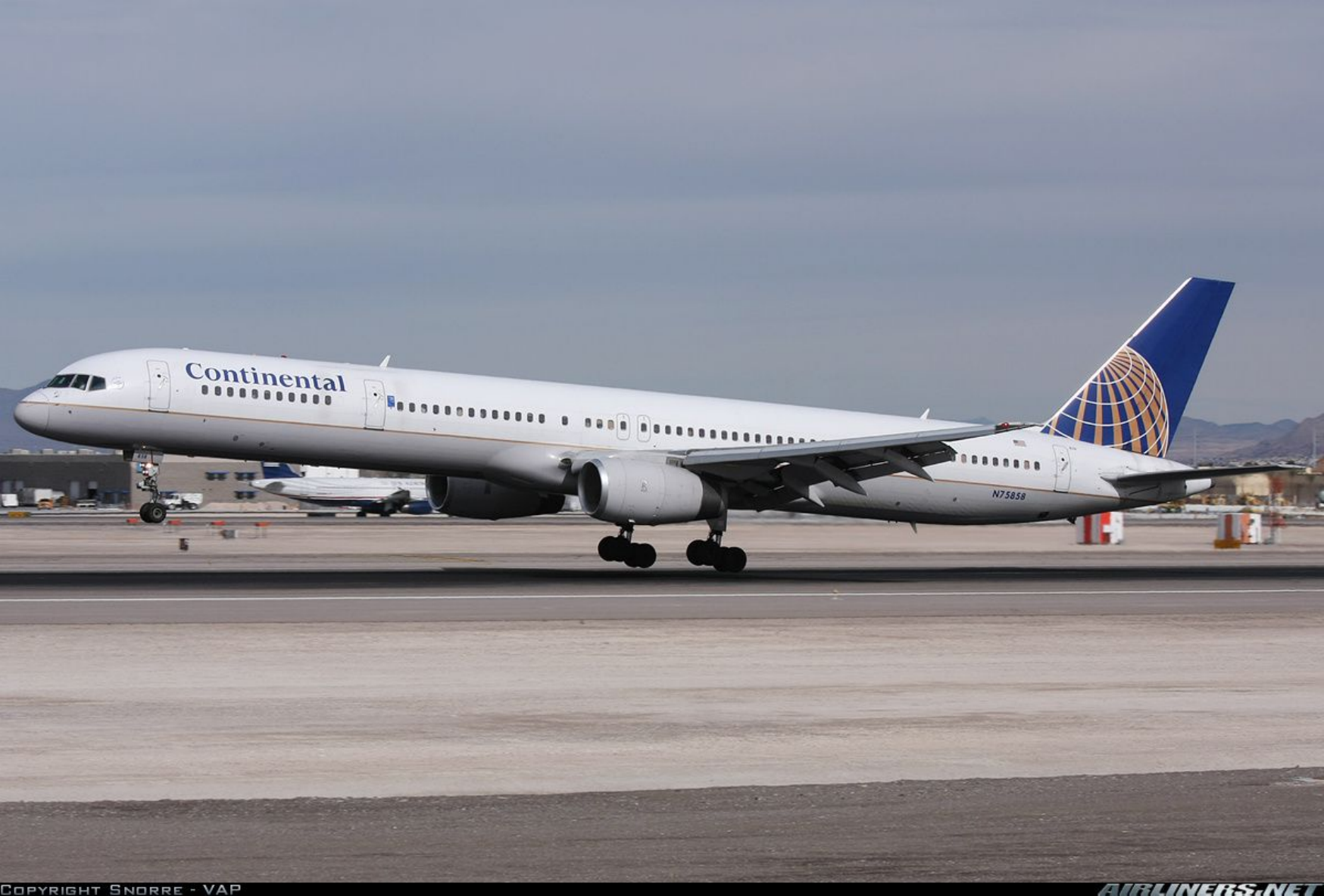}};
        \node at (10.1,-3.65) {\includegraphics[height=0.125\textheight]{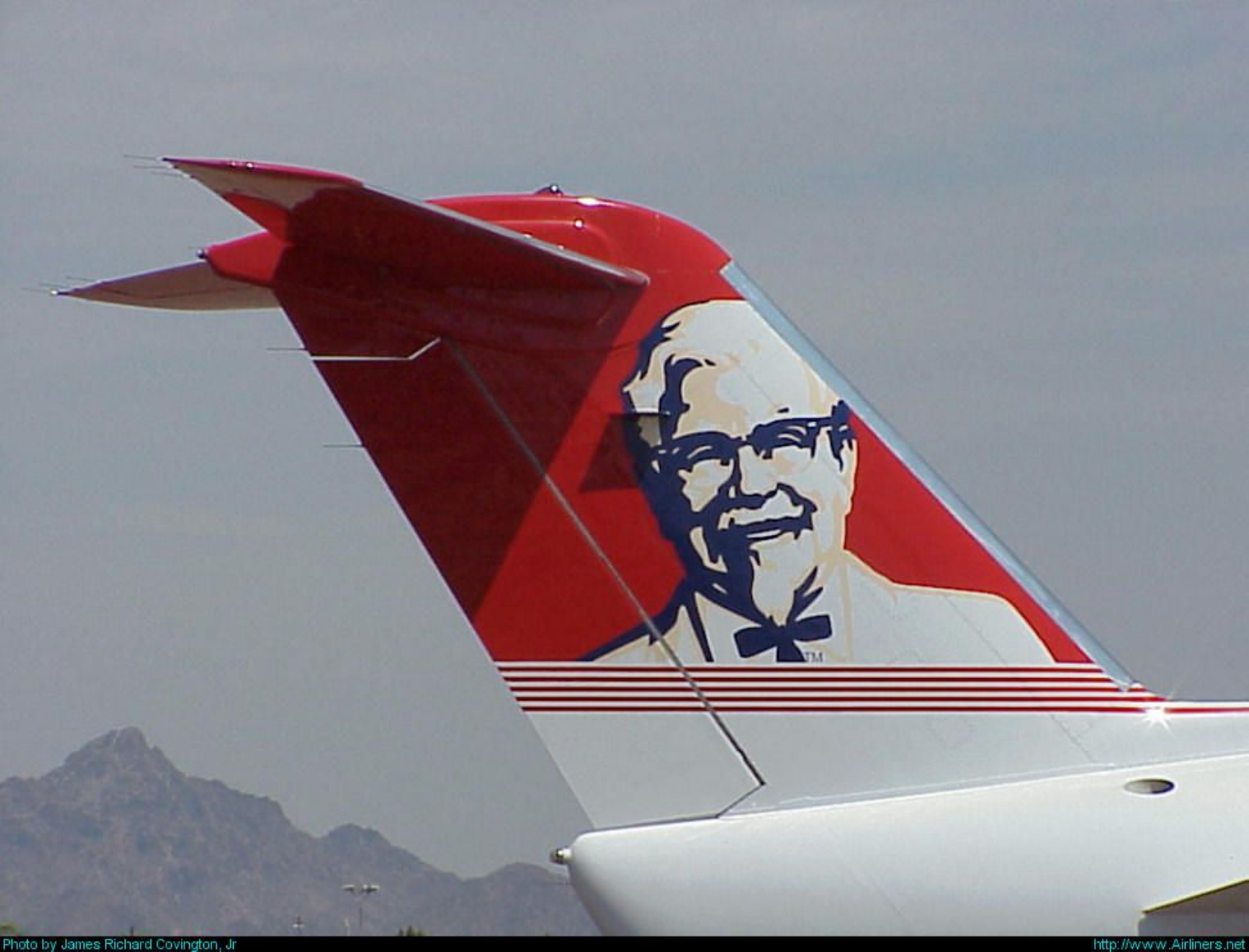}};
    \end{tikzpicture}
    \caption{The histograms on the left illustrate the predictions made over 100 samples with MC-dropout. The column on the right shows the corresponding image. It can be observed that on the first image the model is very certain, whilst on the OOD one (right) it shows almost random guessing.}
    \label{fig:qual_uq}
\end{figure*}

\Cref{fig:qual_uq} presents a qualitative analysis of the model's behavior on in-distribution (ID) and out-of-distribution (OOD) images using 100 Monte Carlo Dropout samples per image. The model's logits are converted to probabilities, and the most likely class is recorded, with histograms illustrating the predicted class distributions. The left histogram shows a tight concentration of predictions for an ID image, indicating high confidence and accuracy in classifying the image. The clear peak reflects the model's strong certainty in this case. The central histogram shows a broader distribution for another ID image, suggesting the model struggles to differentiate between similar sub-classes (e.g., 727-200 vs. 727-300), though it still maintains reasonable confidence. The third histogram shows a dispersed distribution for an OOD image, reflecting significant uncertainty and a lack of confidence. The model appears to be guessing due to incomplete data, as shown by the wide spread of predictions across many classes.

\section{Discussion and Conclusions}

The present paper introduces a novel cross-attention-based prompt tuning approach, which we call Adaptive Prompt Tuning (APT), aimed at enhancing few-shot learning for fine-grained classification.
We evaluated this model, alongside other state-of-the-art prompt tuning approaches, CoOp and VPT, across four datasets: FGVC Aircraft, Oxford Flowers, CUBirds, and Caltech101. The cross-attention mechanism demonstrated significant performance improvements, particularly in datasets with high intra-class variance. For instance, the model achieved substantial gains in the FGVC Aircraft dataset, improving accuracy from 27\% to 47\% as the number of shots increased. However, while these results highlight the approach's efficacy, there remains room for improvement, especially in achieving higher accuracy with fewer training examples and addressing challenges in datasets with complex visual variations.\\

The performance in the Oxford Flowers dataset further highlights the model’s capability, where it reached 97\% accuracy with just 16 shots, though the marginal gains over CoOp in simpler classification tasks suggest that the benefits of cross-attention may be less pronounced in cases with clear and distinct class features. On the CUBirds dataset, the model showed strong results, achieving 77\% accuracy at 16 shots, but the initial performance equaling the zero-shot baseline indicates a need for strategies that can improve early-stage learning. These findings highlight the cross-attention model’s strengths in handling fine-grained distinctions while pointing to opportunities for enhancing performance in fewer-shot scenarios.\\

Uncertainty Quantification (UQ) through Expected Calibration Error (ECE) provided additional insights into the reliability of the uncertainty estimates produced by APT. While the model demonstrated good calibration in the FGVC Aircraft dataset, the higher ECE in comparison to other datasets underscores the impact of accuracy on calibration metrics. The analysis also revealed calibration issues in the CUBirds and Oxford Flowers datasets, where the model was under-confident in lower-confidence predictions and, conversely, overconfident in higher confidence. These discrepancies suggest the need for refined calibration techniques.\\

Future work should focus on improving the model initial learning efficiency and enhancing its calibration. Integrating more advanced data augmentation methods and exploring different configurations of the cross-attention mechanism could further improve the model's performance across varying datasets. Additionally, developing more sophisticated UQ techniques, such as Deep Ensembles or Bayes-by-backprop, might enhance the reliability of uncertainty estimates.


\bibliographystyle{apalike}
{\small
\bibliography{example}}

\appendix
\section{\uppercase{Appendix}}


\subsection{Preliminary Investigations}\label{sec:experiments}

\paragraph{Text Encoder Behaviour}

\begin{figure*}[t]
    \centering
    \includegraphics[width=1\linewidth]{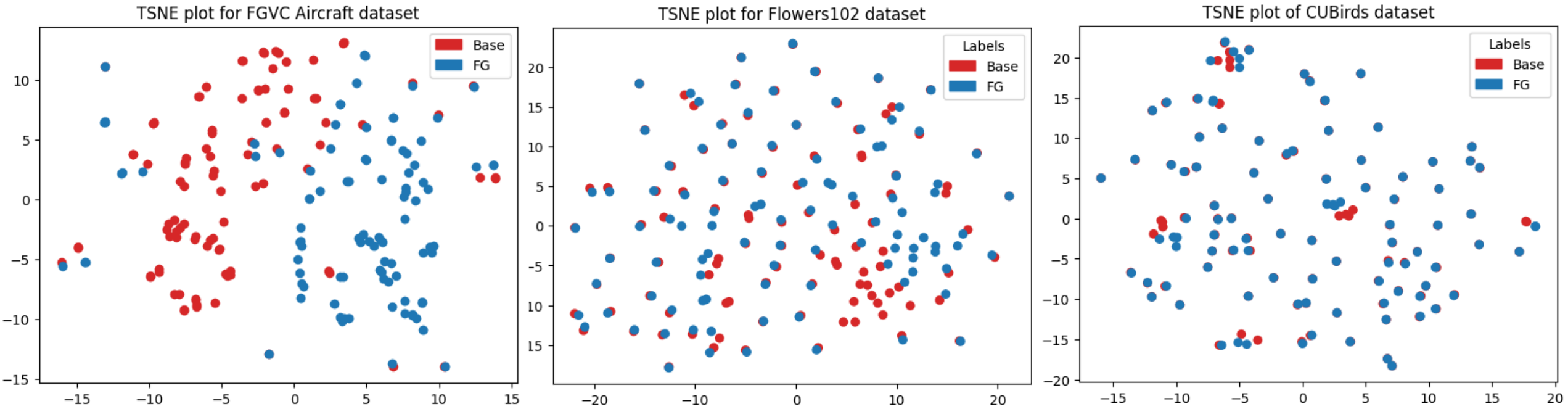}
    \caption{t-SNE plots of baseline prompts (red) versus fine-grained prompts (blue). The FGVC Aircraft dataset shows a larger shift between embeddings, indicating greater influence of surrounding context for less familiar class words.}
    \label{fig:tsne_embed}
\end{figure*}

\begin{table*}[t]
    \centering
    \begin{tabular}{|c|c|c|c|}
    \hline
    & \multicolumn{2}{c|}{\textbf{Image Features} \(\times10^3\)} & \textbf{Text Features} \(\times10^3\) \\
    \hline
       \textbf{Dataset}  & \textbf{Intra-class Variance} & \textbf{Inter-class Variance} & \textbf{CLIP} \\
       \hline
        CUBirds & 0.278 & 0.260 & 0.800  \\
        Oxford Flowers & 0.193 & 0.244 & 0.761  \\
        FGVC Aircraft & 0.371 & 0.219 & 0.401  \\
        \hline
    \end{tabular}
    \caption{Inter-class and intra-class variance for the datasets (values scaled by \( \times10^3 \) for readability). The FGVC Aircraft dataset has higher intra-class variance and lower inter-class variance in image features, and lower inter-class variance in text features, indicating greater classification challenges.}
    \label{tab:class_variances}
\end{table*}

To understand how CLIP's text encoder handles fine-grained versus baseline descriptive texts, we embedded sentences using the text encoder and applied t-SNE for visualization (see Figure~\ref{fig:tsne_embed}). In these plots, baseline prompts are in red, and fine-grained prompts are in blue.

For the CUBirds and Oxford Flowers datasets, there is significant overlap between the embeddings of baseline and fine-grained prompts. This suggests that class names alone are sufficient for capturing semantic features, likely because the model is familiar with these classes from its training data. In contrast, the FGVC Aircraft dataset exhibits a larger divergence between embeddings. The model is less familiar with aircraft class words (e.g., ``A310,'' ``A318''), so additional descriptive context significantly influences the embeddings. These findings imply that static prompts may be inadequate for datasets with uncommon class words, highlighting the need for adaptive prompt-tuning strategies.

\paragraph{Analysis of Variance}

We analyzed inter-class and intra-class variances for the CUBirds, Oxford Flowers, and FGVC Aircraft datasets (see Table~\ref{tab:class_variances}).

In CUBirds and Oxford Flowers, low intra-class variance and higher inter-class variance in image features facilitate classification by creating distinct class clusters with consistent visual representations. This is likely due to structural and color similarities within classes (e.g., plumage patterns, petal structures). Conversely, the FGVC Aircraft dataset exhibits higher intra-class variance and lower inter-class variance, making classification difficult due to greater variability within classes and less distinction between classes. The lower inter-class variance in text features further complicates text-based differentiation. These observations suggest that static prompt-tuning methods like CoOp may be less effective on such complex datasets, emphasizing the necessity for adaptive prompt-tuning methods that can dynamically incorporate visual information.

\subsection{Examples of initial prompting for the text encoder}\label{sec:appendix_text_prompts}

\begin{verbatim}
CUBirds
Baseline:
"A photo of a Blackfooted Albatross,
a type of bird."

Fine-grained: 
"A Blackfooted Albatross, a bird
with long wings, dark head, large body,
rounded tail, stout beak, pale belly."

Oxford Flowers
Baseline:
"A photo of a Pink Primrose, a type of
flower."

Fine-grained: 
"A Pink Primrose, a flower with pink petals,
green sepals, short style, yellow anthers, 
slender filament, cupped receptacle."

FGVC Aircraft
Baseline: 
"A photo of a A310, a type of aircraft."
Fine-grained: 
"A A310, an aircraft with sleek fuselage,
twin engine, narrow wings, T tail, low
cockpit, retractable landing gear"
\end{verbatim}

\end{document}

%% file: figs/diagrams/clip_tikz.tex
\usetikzlibrary{arrows}
\usetikzlibrary{arrows.meta}
\usetikzlibrary{matrix}
\usetikzlibrary{matrix, decorations.pathreplacing}

\begin{tikzpicture}[
            scale=0.725, every node/.style={scale=0.725},
            node distance=0mm,
             box/.style args = {#1/#2}{shape=rectangle,
                    text width=#1mm, minimum height=#2mm,
                    draw, thick, inner sep=0pt, outer sep=0pt, 
                    align=center, text=black}
                ]
    \node (cat) at (-3,0) {\includegraphics[height=2cm]{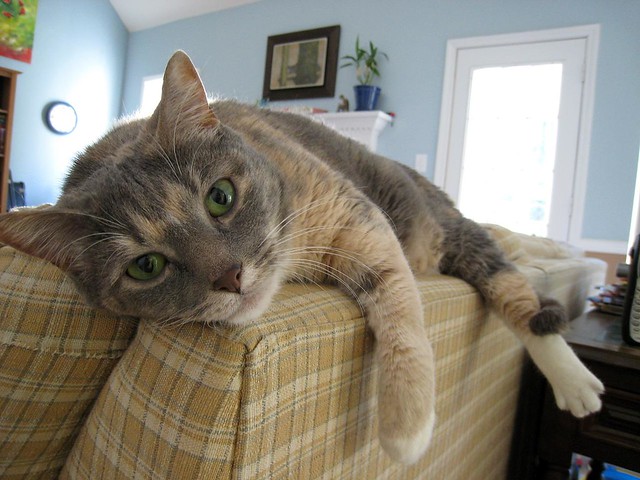}};

    \draw[rounded corners=3pt, fill=yellow] 
        (-1.5, 1) -- (-1.5, -1) -- (0.75, -0.5) -- (0.75, 0.5) -- cycle;
    \node at (-0.335,0.25) {Image encoder};
    \node at (-0.335,-0.25) {\textit{ViT}};

    \node[text width=2cm] (text) at (-3,-2.5) {\texttt{"a cat is relaxing on the back of a plaid couch."}};
    
    \draw[rounded corners=3pt, fill=orange] 
        (-1.5, -1.5) -- (-1.5, -3.5) -- (0.75, -3.0) -- (0.75, -2.0) -- cycle;
    \node at (-0.35, -2.25) {Text encoder};
    \node at (-0.35, -2.75) {\textit{Transformer}};

    \matrix[matrix of nodes, 
            nodes={draw, fill=yellow!60, minimum width=0.5cm, minimum height=0.5cm, anchor=center}, 
            column sep=0pt, row sep=0pt] (image_emb) at (1.5,0) {
        ~ \\
        ~ \\
        ~ \\
        ~ \\
    };
    \draw[decorate, decoration={brace, amplitude=7pt}] (1.8, 1) -- (1.8, -1) node[midway, xshift=18pt] {$\mathbb{R}^d$}; 

    \matrix[matrix of nodes, 
            nodes={draw, fill=orange!60, minimum width=0.5cm, minimum height=0.5cm, anchor=center}, 
            column sep=0pt, row sep=0pt] (image_emb) at (1.5,-2.5) {
        ~ \\
        ~ \\
        ~ \\
        ~ \\
    };
    \draw[decorate, decoration={brace, amplitude=7pt}] (1.8, -1.5) -- (1.8, -3.5) node[midway, xshift=18pt] {$\mathbb{R}^d$};

\end{tikzpicture}

%% file: figs/diagrams/reliabliity.tex
\begin{tikzpicture}
    \begin{axis}[
        axis lines = middle,
        xlabel = {Confidence},
        ylabel = {Accuracy},
        x label style={at={(axis description cs:0.5,-0.05)},anchor=north},
        y label style={at={(axis description cs:-0.05,.5)}, rotate=90, anchor=south},
        xtick={0,0.2,...,1},
        ytick={0,0.2,...,1},
        grid = both,
        domain = 0:1,
        enlarge x limits={abs=0.05},
        enlarge y limits={abs=0.05},
        ymin=0, ymax=1,
        xmin=0, xmax=1,
        ]
        \addplot[domain=0:1, dotted, thick] {x};
        \addplot [
            domain=0:1,
            fill=orange,
            opacity=0.3,
            samples=100,
            forget plot
        ] {x} \closedcycle;

        \addplot [
        fill=yellow,
        domain=0:1,
        samples=2,
        fill opacity=0.5,
        draw=none
    ]
    coordinates {(0,0) (1,1) (0,1) (0,0)};

    \node at (axis cs:0.3, 0.7) {Underconfident model};
    \node at (axis cs:0.65, 0.25) {Overconfident model};
    \node[rotate=41] at (axis cs:0.5, 0.5) {Calibrated model};
    \end{axis}
\end{tikzpicture}